\documentclass[lettersize,journal]{IEEEtran}
\usepackage{amsmath,amsfonts}
\usepackage{algorithmic}
\usepackage{algorithm}
\usepackage{array}
\usepackage{textcomp}
\usepackage{stfloats}
\usepackage{url}
\usepackage{verbatim}
\usepackage{graphicx}
\usepackage{cite}
\usepackage{amssymb}
\usepackage{graphicx}
\usepackage{booktabs}
\usepackage{multirow}
\usepackage{bbm}
\usepackage{amsthm}
\usepackage{bm}
\usepackage{multirow}
\usepackage{rotating}
\usepackage{mathrsfs, dsfont}
\usepackage{stfloats}
\usepackage{bbding}
\usepackage{float}    
\usepackage[caption=false,font=footnotesize]{subfig}

\usepackage{xcolor}
\begin{document}

\title{Integrating SAM Supervision for 3D Weakly Supervised Point Cloud Segmentation}

\author{Lechun You, Zhonghua Wu, Weide Liu, Xulei Yang, Jun Cheng,~\IEEEmembership{Senior Member, IEEE}, Wei Zhou,~\IEEEmembership{Senior Member, IEEE}, Bharadwaj Veeravalli,~\IEEEmembership{Senior Member, IEEE}, Guosheng Lin

\thanks{This research is partly supported by the Agency for Science, Technology and Research (A*STAR) under its MTC Programmatic Funds (Grant No. M23L7b0021).}
\thanks{L. You is with the Department of Electrical and Computer Engineering, National University of Singapore, Singapore 119077 (email: lechun.you@u.nus.edu).}
\thanks{Z. Wu is with SenseTime Research, Singapore 069547 (email: wuzhonghua@sensetime.com).}
\thanks{W. Liu is with the School of Computer Science and Engineering, Nanyang Technological University, Singapore 639798 (email: weide001@e.ntu.edu.sg).}
\thanks{X. Yang and J Cheng are with the Institute for Infocomm Research (I2R), Agency for Science, Technology and Research (A*STAR), Singapore 138632 (email: yang\_xulei@i2r.a-star.edu.sg, cheng\_jun@i2r.a-star.edu.sg).}
\thanks{
W. Zhou is with School of Computer Science and Informatics, Cardiff University, UK (zhouw26@cardiff.ac.uk).
}
\thanks{B. Veeravalli is with the Department of Electrical and Computer Engineering, National University of Singapore, Singapore 119077 (email: elebv@nus.edu.sg).}
\thanks{G. Lin is with the School of Computer Science and Engineering, Nanyang Technological University, Singapore 639798 (email: gslin@ntu.edu.sg).}
\thanks{Manuscript received April 19, 2021; revised August 16, 2021.}}

\markboth{Journal of \LaTeX\ Class Files,~Vol.~14, No.~8, August~2021}%
{Shell \MakeLowercase{\textit{et al.}}: A Sample Article Using IEEEtran.cls for IEEE Journals}

\IEEEpubid{0000--0000/00\$00.00~\copyright~2021 IEEE}

\maketitle

\begin{abstract}
Current methods for 3D semantic segmentation propose training models with limited annotations to address the difficulty of annotating large, irregular, and unordered 3D point cloud data. They usually focus on the 3D domain only, without leveraging the complementary nature of 2D and 3D data. Besides, some methods extend original labels or generate pseudo labels to guide the training, but they often fail to fully use these labels or address the noise within them. Meanwhile, the emergence of comprehensive and adaptable foundation models has offered effective solutions for segmenting 2D data. 
Leveraging this advancement, we present a novel approach that maximizes the utility of sparsely available 3D annotations by incorporating segmentation masks generated by 2D foundation models. We further propagate the 2D segmentation masks into the 3D space by establishing geometric correspondences between 3D scenes and 2D views. We extend the highly sparse annotations to encompass the areas delineated by 3D masks, thereby substantially augmenting the pool of available labels. Furthermore, we apply confidence- and uncertainty-based consistency regularization on augmentations of the 3D point cloud and select the reliable pseudo labels, which are further spread on the 3D masks to generate more labels. This innovative strategy bridges the gap between limited 3D annotations and the powerful capabilities of 2D foundation models, ultimately improving the performance of 3D weakly supervised segmentation.
\end{abstract}


\section{Introduction}

\IEEEPARstart{3}{D} semantic segmentation is a critical computer vision task that assigns semantic labels to each voxel in a three-dimensional scene or each point in a 3D point cloud. This facilitates a thorough comprehension and distinction of various objects or structures in the scene. 3D semantic segmentation has a variety of applications such as autonomous driving, robotics, augmented reality, etc. These techniques can be categorized into three main classes: point-based methods \cite{charles_pointnet_2017,qi_pointnet_2017, li_pointcnn_2018,thomas_kpconv_2019,wu_pointconv_2020,zhao_point_2021,wu_point_2022}, voxel-based methods \cite{graham_3d_2018,choy_4d_2019}, and projection-based methods~\cite{wu_squeezeseg_2018}.

3D point cloud data is a collection of data points obtained by sensors such as LiDAR, RGB-D cameras, laser scanners, etc., with sparse, irregularly sampled, high-dimensional, ambiguous, and unordered characteristics. Due to the intrinsic characteristics of 3D data, annotating extensive data and acquiring high-quality labels poses many challenges. Therefore, many current approaches explore the use of sparse 3D annotations during training \cite{ferrari_3dfeat-net_2018, wei_multi-path_2020, liu_one_2021, wu_dual_2022, tao_seggroup_2022, liu_one_2023, wu_pointmatch_2023, wu_reliability-adaptive_2024}. 

For weakly supervised semantic segmentation tasks, existing research often focuses on 3D point clouds independently, with limited efforts directed toward the integration with 2D images. However, 2D images, known for their simplicity, ease of capture, and detailed textures, naturally enhance the abundance of geometric data offered by 3D data. On the other hand, the convergence of 2D and 3D joint learning primarily revolves around fully supervised semantic segmentation tasks, where features of both modalities are projected and fused \cite{kundu_virtual_2020, alonso_3d-mininet_2020, hu_bidirectional_2021}, or labels are propagated across modalities \cite{genova_learning_2021}, with limited exploration in the domain of weakly supervised tasks.

Additionally, the advances in robust 2D foundational models with zero-shot capabilities have significantly matured the performance of 2D semantic segmentation. Leveraging this progress, we seek to exploit the geometric correspondence between 3D and 2D data, thereby maximizing the utility of sparsely available 3D annotations while still achieving strong performance compared to fully supervised methods.
\IEEEpubidadjcol
We introduce a novel weakly supervised semantic segmentation model designed to address the challenges posed by sparsely annotated 3D point cloud data. This model harnesses the strengths of 2D foundational models in image segmentation. When faced with unlabeled 2D images, we begin by employing the cutting-edge 2D semantic segmentation model, Semantic-SAM \cite{li_semantic-sam_2023}, to produce segmentation masks. We utilize the spatial regions without assigning specific semantic labels, thereby enabling adaptable alignment of class information. Then we leverage geometry information to back-project the 2D masks into the 3D domain and fuse the masks of the same object from different views, yielding 3D masks.

We employ diverse strategies to maximize the utility of limited annotations and 3D masks. Initially, we extend the original sparse annotations onto the 3D masks, significantly increasing the number of available labels. During training, we enforce consistency regularization \cite{wu_reliability-adaptive_2024} to categorize the predictions into reliable and ambiguous subsets and further distribute the reliable pseudo labels across specific regions delineated by the 3D masks in a proportional manner. These expanded labels are essentially accurate, thereby enhancing subsequent training iterations. Acknowledging their potential inadequacy in fully representing each masked region, as well as the presence of noise in the projected masks, we consider them noisy and apply a noise-robust loss \cite{ma_normalized_2020} on them to mitigate this issue.

Our contributions can be summarized as follows.

\begin{itemize}
    \item We present a novel approach for weakly supervised 3D semantic segmentation by utilizing 2D foundation models. By leveraging geometric correspondences between 2D and 3D data, we bridge the gap between 2D foundation models and 3D learning, significantly enhancing label availability by back-projecting 2D segmentation masks into 3D space and extending sparse annotations.
    
    \item We propose a strategy that fuses segmentation masks projected into 3D from different views. This is achieved by evaluating the overlap between different projected masks to determine whether they should be merged into a single mask, ultimately ensuring that each 3D object's mask is complete and non-redundant.

    \item We identify reliable pseudo labels and propose a strategy to refine and expand them using back-projected masks. This strategy is based on the proportion of pseudo labels within each projected mask, optimizing the threshold to achieve maximum accuracy. Our approach significantly increases the number of high-quality labels and enhances overall performance.
    
    \item  Our approach achieves state-of-the-art performance, showcasing its effectiveness in harnessing minimal annotations for 3D data and unlabeled 2D data toward comprehensive 3D scene understanding.
    
\end{itemize}

\section{Related Work}
\subsection{3D Semantic Segmentation}
The research on 3D semantic segmentation is currently widespread, with a focus on leveraging the rich geometric information inherent in 3D data. 

 Charles et al. proposed PointNet \cite{charles_pointnet_2017}, a pioneering work that directly processes irregular raw point cloud data, preserving permutation invariance by applying the symmetric function to aggregate global information. The latter PointNet++ \cite{qi_pointnet_2017} addresses its issue of insufficient capability to capture local information by utilizing a hierarchical structure. Moreover, Zhao et al. introduced the Point Transformer \cite{zhao_point_2021}, which utilizes self-attention networks in 3D point cloud processing. Wu et al. further improved the Point Transformer by introducing group vector attention and position encoding multiplier \cite{wu_point_2022}. Furthermore, the original sample-based pooling is simplified to partition-based pooling. To enhance efficiency and accuracy, they further introduced Point Transformer V3 \cite{wu_point_2024}. This version replaces the traditional neighbor search from K-Nearest Neighbors (KNN) with serialized neighbor mapping, streamlines attention patch interaction mechanisms, and simplifies positional encoding.

Different from handling points directly, Choy et al. \cite{choy_4d_2019} represented the point cloud data as sparse tensors and introduced MinkowskiNet with generalized sparse convolutions for processing 3D sparse voxel grids. Another kind of approach projects the 3D point cloud to multiple 2D views and takes advantage of the capability of 2D CNN and the fine-grained texture of 2D images \cite{wu_squeezeseg_2018}.

Some current methods combine 2D and 3D data in integrated learning schemes. Kundu et al. rendered synthetic 2D images from virtual views of the 3D scene to train a 2D segmentation model, then generated predictive features for fusion on 3D mesh vertices \cite{kundu_virtual_2020}. To enable joint reasoning for 2D and 3D, Hu et al. proposed the Bidirectional Projection Network (BPNet) \cite{hu_bidirectional_2021}, consisting of symmetric 2D and 3D subnets connected at the same decoder level. With a link matrix, 2D and 3D features are correspondingly projected to each other and fused.

However, these methods require fully annotated data to realize their full potential. Yet, annotating point cloud data is resource-intensive due to its unordered and irregular nature.

\subsection{Weakly Supervised 3D Semantic Segmentation}
To address the time-consuming and labor-intensive issue of annotating 3D point cloud data, some methods have proposed training semantic segmentation models on a limited set of labels. Liu et al. introduced the `One-Thing-One-Click (OTOC)' scheme \cite{liu_one_2021}, which means that each object needs only one annotated label. It conducts iterative training and pseudo label propagation, supported by a relation network that learns the similarity among voxels. Nonetheless, the generated pseudo labels are noisy, and the `OTOC' annotation remains costly, as the user must identify every individual object in the scene, with each scene requiring up to 2 minutes of annotation.

Some current research utilizes consistency regularization to weakly supervised point cloud segmentation. Wu et al. introduced the PointMatch \cite{wu_pointmatch_2023} approach that utilizes point-wise predictions from one view as pseudo labels for another view within a point cloud scene, enhancing the overall consistency between the views. Wu et al. proposed RAC-Net \cite{wu_reliability-adaptive_2024}, which leverages both prediction confidence and model uncertainty to divide the pseudo labels into the reliable set and the unreliable set. Then, different consistency constraints are applied to them. These methods directly apply Cross-Entropy loss to pseudo labels and predicted values, but they do not explore further utilization of these pseudo labels.

ActiveST \cite{liu_active_2024} employs an active learning approach for weakly supervised point cloud segmentation tasks. This method automatically selects points with high potential for improving the model, based on prediction uncertainty, for manual annotation, while also using highly confident predictions as pseudo labels for training. However, it requires the user to manually annotate the selected points in each iteration, which distinguishes it from other traditional weakly supervised training methods.

Dong et al. \cite{dong_leveraging_2023} utilized the Segment Anything Model (SAM) \cite{kirillov_segment_2023} to generate segmentation masks for 2D views and propagated sparse 2D labels on these masks. They projected the expanded labels from each view into the 3D scene and employed a voting strategy to aggregate these labels. However, they only utilized the masks in the 2D plane without further exploiting them in the 3D space, and thus missed opportunities to fully leverage the masks' potential in 3D applications. Instead of projecting the expanded 2D labels, we project the 2D segmentation masks to 3D masks to enable flexible use, so that more 3D labels can be expanded on them.

\subsection{2D Foundation Models}
2D semantic segmentation~\cite{liu2020crnet,liu2022few,liu2024harmonizing,liu2025modality} is to label each pixel in an image with a semantic class to provide a detailed understanding of the content of the image. Long et al. proposed Fully Convolutional Networks (FCN) \cite{long_fully_2015} for segmentation and introduced a skip architecture that integrates semantic details from a deep, coarse layer with appearance features from a shallow, fine layer, which has been widely adopted as the backbone in many previous works \cite{ronneberger_u-net_2015, zheng_conditional_2015}. In 2021, Dosovitskiy et al. proposed the Vision Transformer (ViT) \cite{dosovitskiy_image_2021}, which utilizes self-attention mechanisms. It divides an image into fixed-size patches and preserves spatial information through positional embeddings. ViT exhibits strong scalability and high performance, which makes it widely adopted in various image-processing tasks. Currently, many methods also improve upon ViT for 2D image segmentation \cite{zhang_segvit_2022, liu_swin_2021}.

In recent times, with the rapid advancement of Large Language Models (LLM), some studies have integrated open vocabulary into visual tasks to establish foundational models for image segmentation. They seek to divide an image into semantic regions based on arbitrary text descriptions. Kirillov et al. proposed the Segment Anything Model (SAM) \cite{kirillov_segment_2023}, which supports any segmentation prompt with zero-shot generalization. SAM3D \cite{yang_sam3d_2023} utilized SAM to generate masks for RGB images and then projected these masks onto 3D point clouds. By iteratively merging the masks, they achieved fine-grained segmentation results. However, this method strictly requires 2D views that are matched with the 3D scene. When inferring new data, images need to be re-collected for the new data. However, our approach only utilizes 2D images during the training phase to alleviate the annotation burden, while during inference, it directly uses 3D point clouds.

Building upon SAM, Li et al. introduced Semantic-SAM \cite{li_semantic-sam_2023}, which further enables image segmentation at any given granularity by jointly training the model on seven datasets with multiple semantic labels. Wang et al. introduced a hierarchical representation approach encompassing semantic, instance, and part levels \cite{wang_hierarchical_2023}. This approach decouples the representation learning modules and text-image fusion mechanisms for both background and foreground.

2D foundation models, trained on large-scale 2D datasets and incorporating language modalities, demonstrate excellent generalization capabilities and zero-shot performance. This makes them ideal complements to 3D data with the inherent geometric correspondences. We opt for the Semantic-SAM \cite{li_semantic-sam_2023} to generate 2D semantic segmentation masks due to its strong performance and flexible granularity options, including semantic, instance, and part levels.

\subsection{Learning From Noisy Labels}
Several techniques have been proposed to address the challenge of training accurate deep-learning models in the presence of noisy labels. An approach to mitigate the impact of noisy labels is to use robust loss functions, which offer simplicity in implementation. Ghosh et al. theoretically demonstrated that loss functions such as the Mean Absolute Error (MAE) exhibit noise-robust properties, whereas the commonly employed Cross-Entropy (CE) loss function does not possess such robustness \cite{ghosh_robust_2017}. Zhang et al. introduced the Generalized Cross-Entropy (GCE) loss \cite{zhang_generalized_2018}, which can be readily applied to existing networks. Furthermore, Wang et al. augmented the original CE loss with a Reverse Cross-Entropy (RCE) term, forming the Symmetric Cross-Entropy (SCE) loss \cite{wang_symmetric_2019}. Furthermore, Ma et al. demonstrated that normalization could render any loss robust to noisy labels \cite{ma_normalized_2020}. They compared the performance of models trained with different combinations of normalized loss functions. Zhou et al. rectified commonly used loss functions and proposed asymmetric loss functions to deal with multiple types of noise \cite{zhou_asymmetric_2021}.

In addition to utilizing robust loss functions, other methodologies focus on detecting and rectifying mislabeled data points \cite{arazo_unsupervised_2019}. For the point cloud segmentation task, Ye et al. proposed a confidence-based approach to select reliable labels, using historical predictions for each data point, and employing a voting strategy to generate labels within clusters \cite{ye_learning_2021}.

\section{Method}
\label{method}
\begin{figure*}[htbp]
    \centering
    \includegraphics[width=1\linewidth]{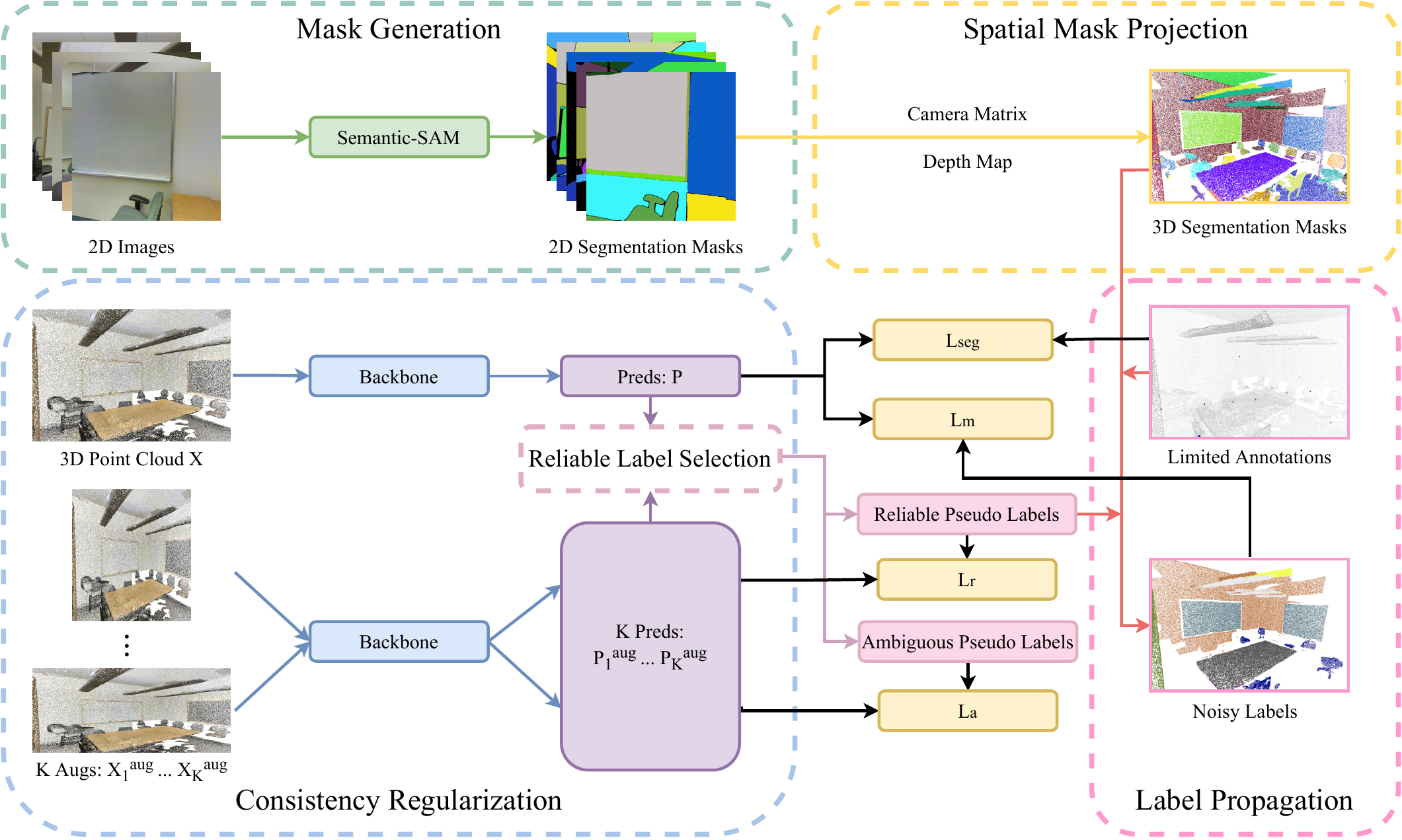}
    \caption{Our method consists of four main components. The Mask Generation and Spatial Mask Projection modules are to generate segmentation masks for 2D views and project them into 3D space, as discussed in Section \ref{mask}. The Consistency Regularization module outlines the process of obtaining reliable pseudo labels and is detailed in Section \ref{rac}. The Label Propagation module is responsible for spreading limited annotations and reliable pseudo labels to masked regions, as elaborated in Section \ref{init}. }
    \label{mine}
\end{figure*}

Our weakly supervised setting is defined as follows: during training, the required data includes 3D point clouds $X$, where only a small subset of points are annotated, referred to as limited annotations $Y$, while the rest remain unlabeled, along with RGB images that correspond to each point cloud. During inference, only the 3D point cloud is required.

The entire framework is illustrated in Figure \ref{mine}.  First, Semantic-SAM \cite{li_semantic-sam_2023} is used to generate 2D segmentation masks for each view within a scene. These masks accurately outline pixel clusters corresponding to each object. Subsequently, leveraging the spatial correspondence between 2D views and the 3D scene, these 2D masks are back-projected onto the 3D space, discussed in Section \ref{mask}. We explore various strategies to maximize the utility of the 3D masks. First, we extend the limited 3D annotations onto the masks, creating additional training labels. But they are still around the initial annotations, failing to cover all the masked regions. To make full use of the masks, we select reliable pseudo labels and propagate them onto 3D masks to expand their coverage during training. They are further fused with the initial expanded annotations, discussed in Section \ref{init}. However, despite the increase in the number of labels, noise is introduced. To mitigate this issue, a noise-robust normalized loss \cite{ma_normalized_2020} is implemented to train the model with the noisy labels, detailed in Section \ref{all}. To classify the pseudo labels into reliable and ambiguous subsets, consistency regularization is employed based on prediction confidence and uncertainty \cite{wu_reliability-adaptive_2024}, detailed in Section \ref{rac}. 


\subsection{Spatial Mask Projection}
\label{mask}
To harness the robust capabilities of foundation models for optimizing the utilization of limited annotations in our segmentation tasks, the Semantic-SAM \cite{li_semantic-sam_2023} is employed to generate segmentation masks for 2D views. We chose Semantic-SAM because it has zero-shot capabilities, supports flexible segmentation at six different granularities, and allows for combining different granularities to generate segmentation masks.

For all 2D views corresponding to each 3D scene, adjacent views often have significant overlap, with the same object being captured in multiple frames. To save runtime and computational resources, we uniformly sample $N_{view}$ views from all available 2D views to ensure that the captured objects are as complete as possible. 

For each sampled 2D view, the set of segmentation masks is represented as a boolean matrix $Mask_{2D}$ of dimensions $[M, H, W]$, where $M$ denotes the number of classes, $H$ and $W$ represent the height and width of the image. There are $M$ boolean values corresponding to each pixel, with only one being true, indicating that the pixel is assigned to the class represented by that index. 

After that, $Mask_{2D}$ are propagated to the 3D domain using the spatial correspondence between 2D pixels and 3D points through a link matrix $\mathcal{L}$ \cite{hu_bidirectional_2021}. The initial step is to compute matrix $\mathcal{M}$, the multiplication of the intrinsic camera calibration matrix $\mathbf{K}$ by the extrinsic camera pose matrix $[\mathbf{R} \vert \mathbf{t}]$ consisting of rotation $\mathbf{R}$ and translation $\mathbf{t}$, denoted as:
\begin{equation}
    \mathcal{M} = \mathbf{K}[\mathbf{R} \vert \mathbf{t}]
\end{equation}

Therefore, the projection from 3D homogeneous coordinates $[x_i, y_i, z_i, 1]^T$ to 2D homogeneous coordinates $[u_i, v_i, 1]^T$ corresponding to the $i^{th}$ 3D point can be denoted as:
\begin{equation}
    [u_i, v_i, 1]^T = \mathcal{M}[x_i, y_i, z_i, 1]^T
\end{equation}

The treatment of occlusion is addressed by incorporating depth information. Finally, $\mathcal{L}$ is an $N \times 3$ matrix which reveals correspondences between 2D and 3D points:
\begin{equation}
    \mathcal{L}_i= [u_i, v_i, m_i],\\
\end{equation}
\begin{equation}
    m_i=
    \begin{cases}
        1,& \text{ if } U_{min} \leq u_i \leq U_{max} \\
        & \text{ and } V_{min} \leq v_i \leq V_{max} \\
        & \text{ and } \vert d(u_i, v_i)-z_i'\vert \leq \delta\\
        0,& \text{ otherwise}
    \end{cases}
\end{equation}
where $N$ is the number of 3D points; $m_i$ is the binary mask reveals whether the $i^{th}$ 3D point has the valid projected 2D pixel; $U_{min}$, $U_{max}$, $V_{min}$, $V_{max}$ are the boundaries of the view frustum; $d(\cdot)$ is the mapping from coordinates to the depth; $z_i'$ is the projected $z$ coordinate of the point; $\delta$ is the threshold for depth matching.

For each 2D view, its segmentation masks $Mask_{2D}$ are projected onto the 3D masks $Mask'_{3D}$ based on the above correspondence. Therefore, the projected matrix $Mask'_{3D}$ has the dimensions of $[M, N]$, calculated by:
\begin{equation}
    Mask'_{3D}(i) =
    \begin{cases}
    Mask_{2D}(u_i, v_i)_{M \times 1}, & \text{if } m_i = 1 \\
    \mathbf{False}_{M \times 1}, & \text{if } m_i = 0
    \end{cases}
\end{equation}

As a result, we obtained $N_{view}$ sets of 3D masks, each corresponding to the masks from different 2D views. These 3D masks are then merged to handle overlapping regions. The merging process works as follows: we initialize $Mask_{3D}$ with the first set of 3D masks projected from the first view, and then sequentially examine the subsequent sets of 3D masks. If a mask from the current set (current view) overlaps with a mask from the previous set (previous view) by more than a certain threshold, we assume they belong to the same class and merge the two masks. Otherwise, the current mask is added as a new mask. This process continues until all masks have been processed. The final result is a matrix $Mask_{3D}$ with dimensions $[T, N]$, where $T$ is the number of object classes after merging. Each 3D point is assigned $T$ boolean values, indicating whether it belongs to one of the $T$ classes.

\subsection{Consistency Regularization and Reliable Pseudo Label Selection}
\label{rac}
Motivated by the methodology of RAC-Net \cite{wu_reliability-adaptive_2024}, we seek to enhance the utilization of limited annotations and select reliable pseudo labels through the incorporation of consistency regularization.

We feed the original point cloud $X$ into the backbone network (which can be any point cloud segmentation network), alongside its $K$ augmented counterparts $X^{aug}_{1...K}$, to obtain $P$ and $P^{aug}_{1...K}$. The augmentation methods we use are PointWolf \cite{kim_point_2021} and Affine Transformation (AT).

\begin{equation}
    \begin{aligned}
        P &= \text{backbone}(X) \\
        P_{1}^{aug} &= \text{backbone}(X_{1}^{aug}) \\
        &... \\
        P_{K}^{aug} &= \text{backbone}(X_{K}^{aug})
    \end{aligned}
\end{equation}

Subsequently, following the procedure employed in the RAC-Net, the predictions are categorized into a reliable set $P^r$ and an ambiguous set $P^a$:
\begin{equation}
    \begin{aligned}
        P^r &= R \cdot P, \quad P^a = (1 - R) \cdot P, \\
        R &= \mathds{1} \sum^C_{c=1}(\mathds{1}[\bar{P}_c \geq \tau] \cdot \mathds{1}[\sigma(\hat{P}_c) \leq \kappa]) > 0
    \end{aligned}
\end{equation}

where $\bar{P}_c$ denotes the confidence of predictions, calculated from the mean of predictions for both the original and augmented data; $\sigma(\hat{P}_c)$ signifies the uncertainty of the predictions, which is the statistical variance; $\mathds{1}$ is the indicator function. For the prediction of each point, if both conditions are satisfied — the prediction confidence $\bar{P}_c$ is greater than or equal to the threshold $\tau$ and the uncertainty $\sigma(\hat{P}_c)$ is less than or equal to the threshold $\kappa$ — then the prediction is considered reliable; otherwise, it is considered unreliable.

The reliable predictions $P^r$ are transformed into one-hot pseudo labels $\widetilde{Y^r}$. Then, the Cross-Entropy loss is applied between $\widetilde{Y^r}$ and predictions of augmented data:
\begin{equation}
\begin{aligned}
    \mathbf{L_r} &= CE[\widetilde{Y^r}, R \cdot P^{aug}_{1}] \\
    & + \ldots + CE[\widetilde{Y^r}, R \cdot P^{aug}_{K}]
\end{aligned}
\end{equation}

For the ambiguous predictions $P^a$, the KL Divergence is computed between the soft pseudo labels $P^a$ and the predictions of augmented data:
\begin{equation}
    \begin{aligned}
        \mathbf{L_a} & = KL[P^a, (1-R) \cdot P^{aug}_{1}] \\
        & + \ldots + KL[P^a, (1-R) \cdot P^{aug}_{K}]
    \end{aligned}
\end{equation}

\subsection{Label Propagation}
\label{init}
In this module, we propagate both the limited annotations $Y$ and reliable pseudo labels $\widetilde{Y^r}$ onto the masks, which significantly increases the number of available labels.

\textbf{Label Initialization.} Given the constraint of very limited labels $Y$, more labels are produced by propagating them onto specific regions indicated by $Mask_{3D}$. For each region in $Mask_{3D}$, we simply count the annotations and distribute their mode within the region.

\textbf{Expansion of Reliable Pseudo Labels.} The reliable pseudo labels $\widetilde{Y^r}$ are also propagated to $Mask_{3D}$ by selecting the mode of them in each region specified by each mask. If the proportion of $\widetilde{Y^r}$ that equals the mode $label_{m}$ exceeds a certain threshold $\eta$, they are considered as the labels for all points in the region. Here,  $\eta$ is a hyperparameter that determines whether the mode of reliable pseudo labels is expanded to the entire mask.

Finally, the expanded reliable pseudo labels are fused with the expanded annotations, denoted as $\widetilde{Y}$. This procedure is shown in Algorithm \ref{prop}.

\begin{algorithm}[H]
\caption{Label Propagation}
\label{prop}
\begin{algorithmic}
\STATE 
\STATE {\textsc{PROPAGATE}}$(Y, \widetilde{Y^r}, Mask_{3D}, T, \eta)$
\STATE \hspace{0.5cm} \textbf{for } $t = 1$ to $T$ \textbf{ do}
\STATE \hspace{1.0cm} $mask \gets Mask_{3D}(t)$
\STATE \hspace{1.0cm} $label_m \gets mode(\widetilde{Y^r} \cdot mask)$
\STATE \hspace{1.0cm} \textbf{if } 
$ \displaystyle \frac{\left| (\widetilde{Y^r}\cdot mask) = label_m \right|}{\left| mask \right|} > \eta$
\textbf{ then}
\STATE \hspace{1.5cm} $(\widetilde{Y} \cdot mask) \gets label_m$
\STATE \hspace{1.0cm} \textbf{else if} $not\_empty(Y \cdot mask)$ \textbf{ then}
\STATE \hspace{1.5cm} $(\widetilde{Y} \cdot mask) \gets mode(Y \cdot mask)$
\STATE 
\STATE \textbf{return} $\widetilde{Y}$
\end{algorithmic}
\end{algorithm}

Even with the expansion of the available labels, they remain susceptible to inaccuracies or incompleteness because of misalignments among categories in the 2D segmentation mask, inaccuracies in the 2D-3D projection, or points falling on boundaries and being erroneously propagated to nearby masks. This problem can be effectively tackled by implementing robust learning methods specifically designed to handle noisy labels.

\subsection{Overall Loss Function}
\label{all}

The propagated labels $\widetilde{Y}$ are treated as noisy labels, and the normalized loss \cite{ma_normalized_2020} is applied to them, which are Normalized Cross-Entropy (NCE) loss and Reverse Cross-Entropy (RCE) loss:
\begin{equation}
    \mathbf{L_m} = NCE[\widetilde{Y}, P] + RCE[\widetilde{Y}, P],
\end{equation}
where $P$ is the prediction of original point clouds $X$, and NCE and RCE are defined as:
\begin{equation} 
    \begin{aligned}
        NCE & = \frac{-\sum^K_{k=1}\mathbf{q}(k\vert x)log\mathbf{p}(k\vert x)}{-\sum^K_{j=1}\sum^K_{k=1}\mathbf{q}(y=j\vert x)log\mathbf{p}(k\vert x)}, \\
        RCE & = -\sum^K_{k=1}\mathbf{p}(k\vert x)log\mathbf{q}(k\vert x),
    \end{aligned}
\end{equation}
where $y$ is the label of input $x$ in the $K$-class classification problem, $\mathbf{p}(k\vert x)$ is the softmax prediction, $\mathbf{q}(k\vert x)$ represents the distribution over different labels for sampler.

Furthermore, the Cross-Entropy (CE) loss is applied to the prediction $P$ and the original sparse labels $Y$.
\begin{equation}
    \mathbf{L_{seg}} = CE[Y, P]
\end{equation}

Finally, the overall loss function is the weighted sum of the above losses:
\begin{equation}
    \mathbf{L} = \lambda_{seg}\mathbf{L_{seg}} + \lambda_{r}\mathbf{L_r} + \lambda_{a}\mathbf{L_a} + \lambda_{m}\mathbf{L_m}
\end{equation}

\section{Experiment}
\subsection{Experimental Setup}

The experiments were conducted using two datasets: ScanNetV2 \cite{dai_scannet_2017} and S3DIS \cite{armeni_joint_2017}. ScanNetV2 comprises 1201 training and 312 validation scans sourced from 706 distinct scenes, with an additional test set of 100 scans. The experiments focus on the `3D Semantic Label with Limited Annotations' benchmark, employing 20 training points per scene.

The S3DIS dataset includes 3D scans of 271 rooms across 6 distinct areas, each containing 13 categories. For training, data from Areas 1, 2, 3, 4, and 6 are utilized, and the model's performance is evaluated on Area 5. Since there's no official limited annotation setting provided, the `One-Thing-One-Click (OTOC)' annotation scheme \cite{liu_one_2021} is adopted by randomly retaining one annotation in each instance.

The evaluation metrics employed in this study involve the calculation of the mean of class-wise intersection over union (mIoU).

\subsection{Comparison on the ScanNetV2 and S3DIS dataset}
Table \ref{scannet3d} shows the mIoU results of previous methods and our method on the ScanNetV2 3D semantic label benchmark. Table \ref{s3dis} presents the mIoU results obtained by previous methods alongside the proposed approach on the S3DIS 3D semantic segmentation dataset, with evaluations conducted on Area 5 as the testing set. For the two tables, the first part includes pure 3D data-based fully supervised methods, the second part shows fully supervised methods that leverage 2D information, and the third part includes weakly supervised methods based on 3D data. Our experiments are conducted on Point Transformer V3 (PTv3) \cite{wu_point_2024} backbone, and results are shown in the fourth part. 

On both datasets, our approach surpasses the baseline by more than 9\%, outperforms some fully supervised methods, and achieves state-of-the-art performance among weakly supervised methods, demonstrating its effectiveness.

\begin{table}[htbp]
    \centering
    \caption{Comparison on ScanNetV2 testing set. Our method achieves the top performance among weakly supervised methods, with 20 training points per scene.}
    \begin{tabular}{ccc}
        \toprule
        Method & Supervision  & mIoU(\%)   \\
        \midrule
        PointNet++ \cite{qi_pointnet_2017}    & 100\%  & 33.9  \\
        PointCNN \cite{li_pointcnn_2018}  & 100\%  & 45.8 \\
        SparseConvNet \cite{graham_3d_2018}  & 100\%  & 72.5 \\
        KPConv \cite{thomas_kpconv_2019} & 100\%  & 68.6 \\
        MinkowskiNet \cite{choy_4d_2019} & 100\%  & 73.6 \\
        PointConv \cite{wu_pointconv_2020} & 100\%  & 66.6 \\
        Point Transformer V2 \cite{wu_point_2022} & 100\%  & 75.2 \\
        \midrule
        3DMV \cite{dai_3dmv_2018} & 100\%  & 48.4 \\
        Virtual MVFusion \cite{kundu_virtual_2020}  & 100\%  & 74.6 \\
        BPNet \cite{hu_bidirectional_2021} & 100\%  & 74.9 \\
        \midrule
        PointContrast\_LA\_SEM \cite{xie_pointcontrast_2020} & 20 points & 55.0 \\
        Viewpoint\_BN\_LA\_AIR \cite{luo_pointly-supervised_2021} & 20 points & 54.8 \\
        One-Thing-One-Click \cite{liu_one_2021}& 20 points  & 59.4 \\
        PointMatch \cite{wu_pointmatch_2023} & 20 points  & 62.4 \\
        RAC-Net \cite{wu_reliability-adaptive_2024} & 20 points &63.9 \\
        \midrule
        Our Baseline (PTv3) & 20 points & 60.1\\
        \textbf{Ours} & 20 points & \textbf{69.4} \\
        Our Upper Bound (PTv3) & 100\%  & 77.9\\
        \bottomrule
    \end{tabular}
    \label{scannet3d}
\end{table}
\begin{table}[htbp]
    \centering
    \caption{Comparison on S3DIS testing set (Area 5). Our method achieves the best results among weakly supervised methods under the `OTOC' setting.}
    \begin{tabular}{ccc}
        \toprule
        Method & Supervision  & mIoU(\%)   \\
        \midrule
        PointNet \cite{charles_pointnet_2017}    & 100\%  & 41.1  \\
        PointCNN \cite{li_pointcnn_2018}  & 100\%  & 57.3 \\
        KPConv \cite{thomas_kpconv_2019} & 100\%  & 65.4 \\
        MinkowskiNet \cite{choy_4d_2019} & 100\%  & 65.4 \\
        Point Transformer V2 \cite{wu_point_2022} & 100\%  & 71.6 \\
        \midrule
        Virtual MVFusion \cite{kundu_virtual_2020}  & 100\%  & 65.4 \\
        \midrule
        One-Thing-One-Click \cite{liu_one_2021}& 0.02\%(OTOC)  & 50.1 \\
        PointMatch \cite{wu_pointmatch_2023} & 0.02\%(OTOC)  & 55.3 \\
        RAC-Net \cite{wu_reliability-adaptive_2024} & 0.02\%(OTOC) & 58.4 \\
        \midrule
        Our Baseline (PTv3) & 0.02\%(OTOC) & 54.7\\
        \textbf{Ours} & 0.02\%(OTOC) & \textbf{64.4} \\
        Our Upper Bound (PTv3) & 100\%  & 73.4 \\
        \bottomrule
    \end{tabular}
    
    \label{s3dis}
\end{table}
Our method can be applied with any point cloud segmentation backbone. We conducted experiments using the ScanNetV2 dataset with the SparseUNet \cite{spconv_contributors_spconv_2022} backbone to validate its effectiveness. Table \ref{scannet_sp} presents the results of the ScanNet V2 validation set.
\begin{table}[htbp]
    \centering
    \caption{Comparison on ScanNetV2 validation set with SparseUNet \cite{spconv_contributors_spconv_2022} backbone.}
    \begin{tabular}{ccc}
        \toprule
        Method & Supervision  & mIoU(\%)   \\
        \midrule
        Our Baseline (SparseUNet) & 20 points & 59.2 \\
        \textbf{Ours} & 20 points & \textbf{67.4} \\
        Our Upper Bound (SparseUNet) & 100\%  & 74.8 \\
        \bottomrule
    \end{tabular}
    
    \label{scannet_sp}
\end{table}

\subsection{Discussions}
\subsubsection{Effectiveness of Back-Projected Masks}
This section explores the effectiveness of back-projecting 2D segmentation masks to 3D masks and propagating the limited annotations $Y$ onto the masks, by comparing the number and accuracy of expanded labels and training the model using the expanded labels.

\begin{figure*}[hb]
	\centering
        \captionsetup[subfigure]{labelformat=empty}
        \subfloat{\includegraphics[width=.2\linewidth]{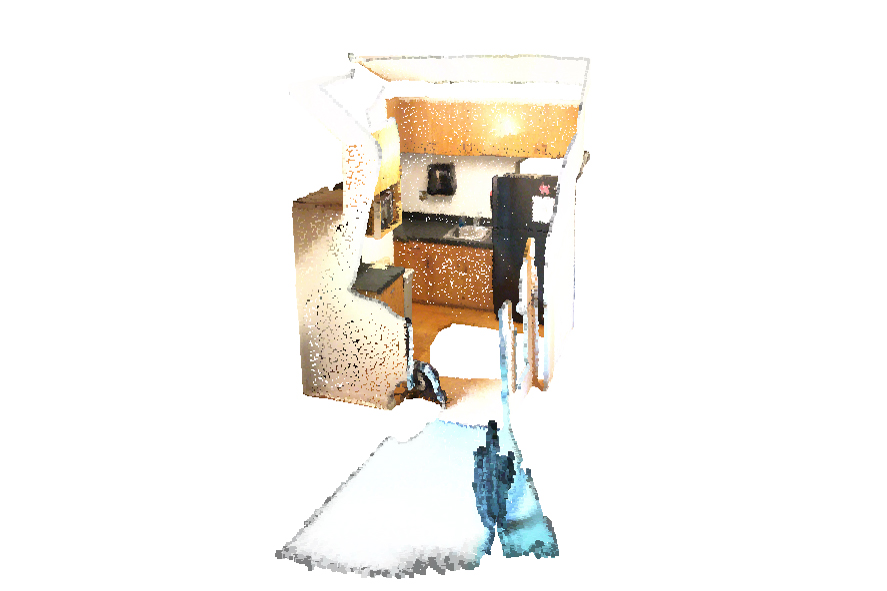}}
	\subfloat{\includegraphics[width=.2\linewidth]{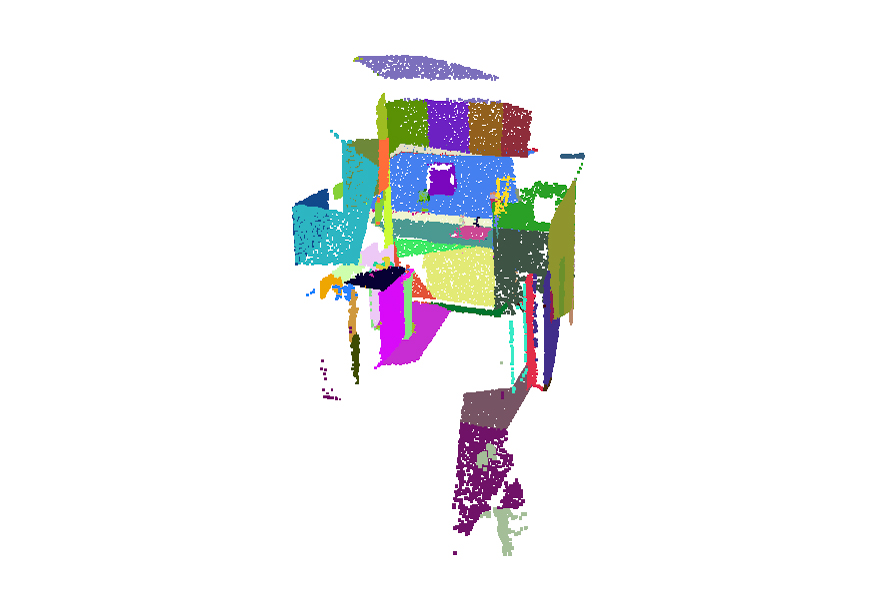}}
        \subfloat{\includegraphics[width=.2\linewidth]{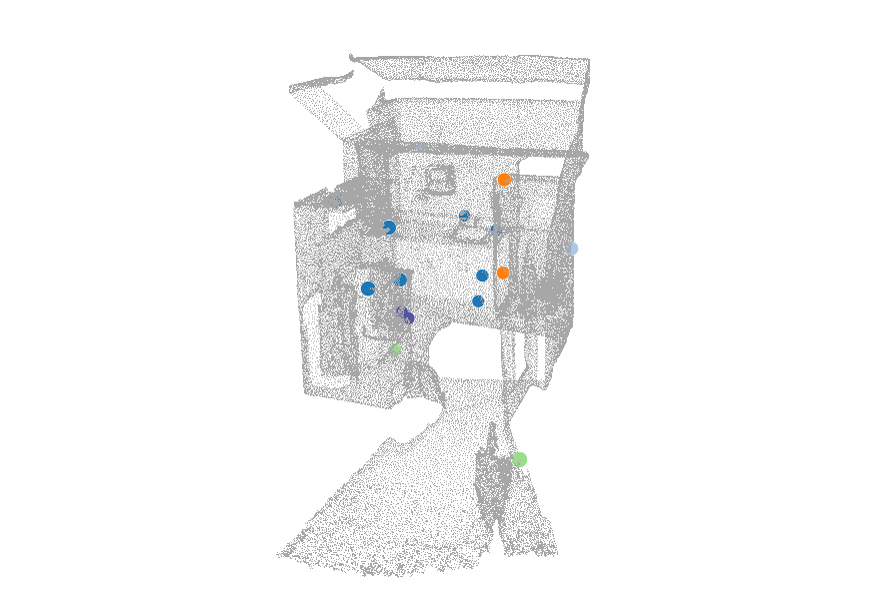}}
        \subfloat{\includegraphics[width=.2\linewidth]{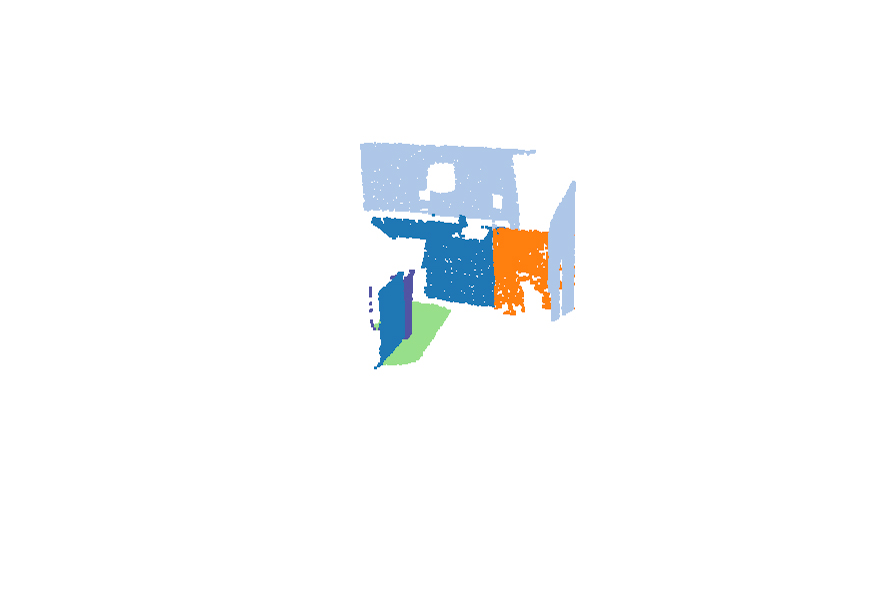}}
        \subfloat{\includegraphics[width=.2\linewidth]{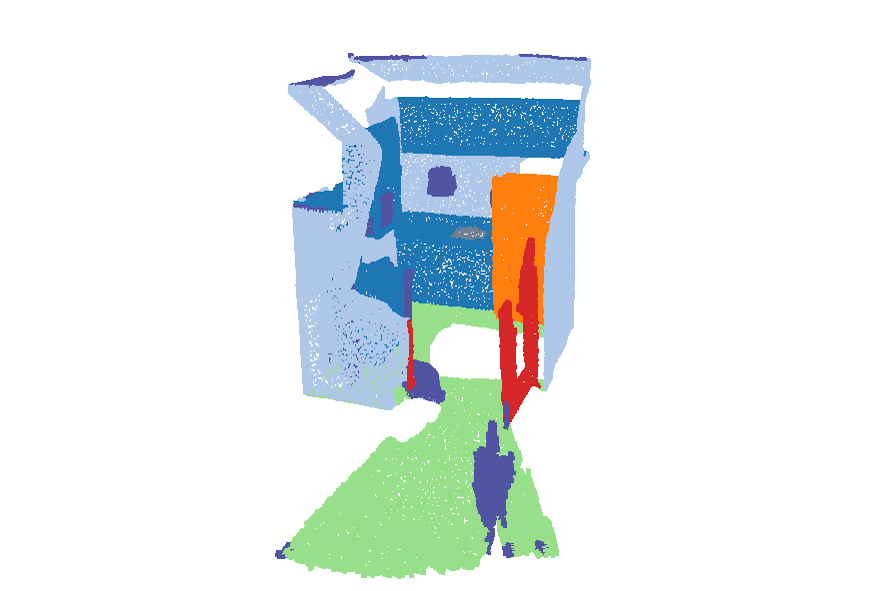}} \\
	\subfloat{\includegraphics[width=.2\linewidth]{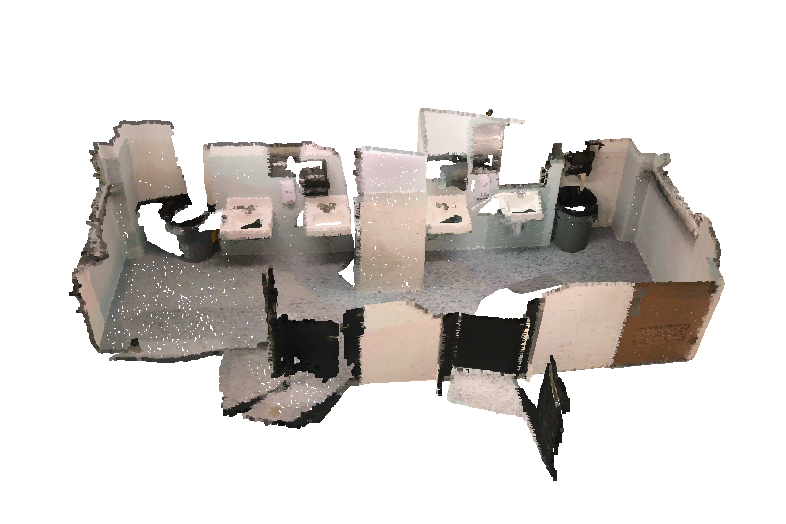}}
	\subfloat{\includegraphics[width=.2\linewidth]{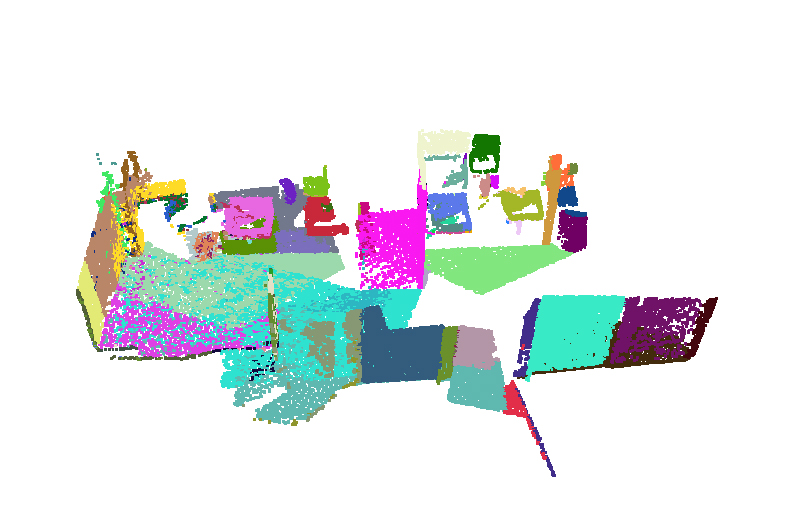}}
        \subfloat{\includegraphics[width=.2\linewidth]{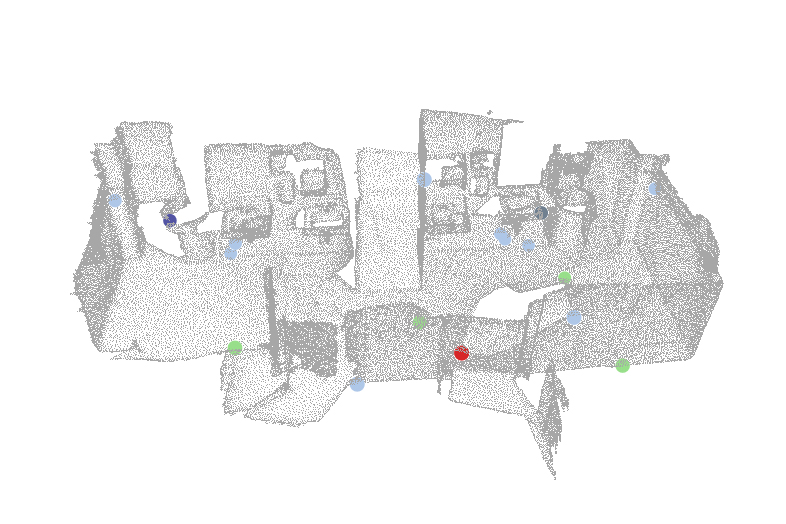}}
        \subfloat{\includegraphics[width=.2\linewidth]{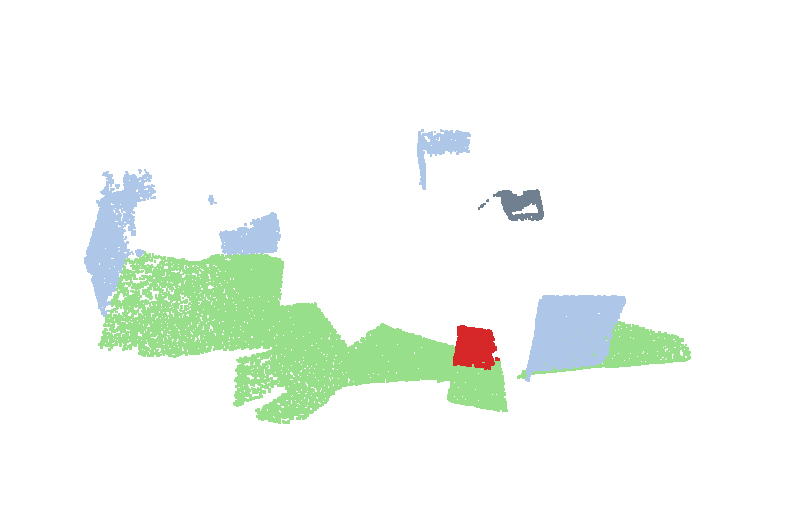}}
        \subfloat{\includegraphics[width=.2\linewidth]{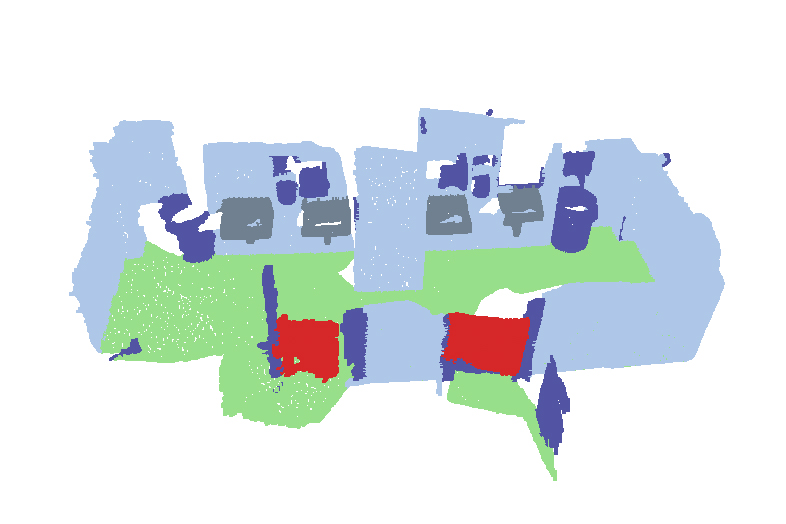}} \\
	\subfloat[Point Clouds]{\includegraphics[width=.2\linewidth]{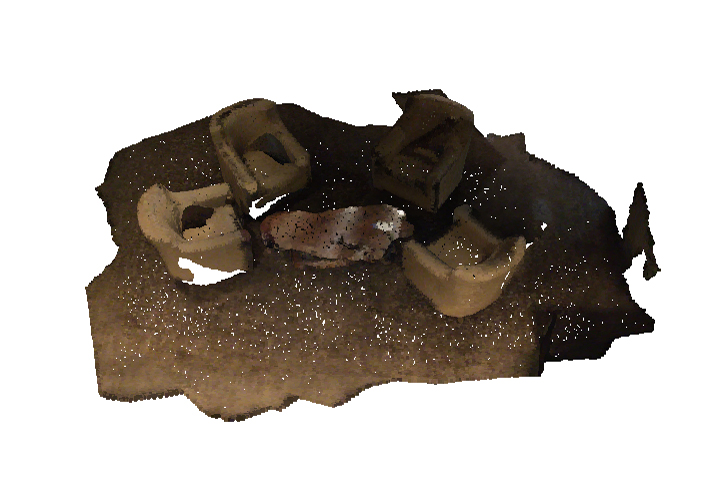}}
	\subfloat[3D Masks]{\includegraphics[width=.2\linewidth]{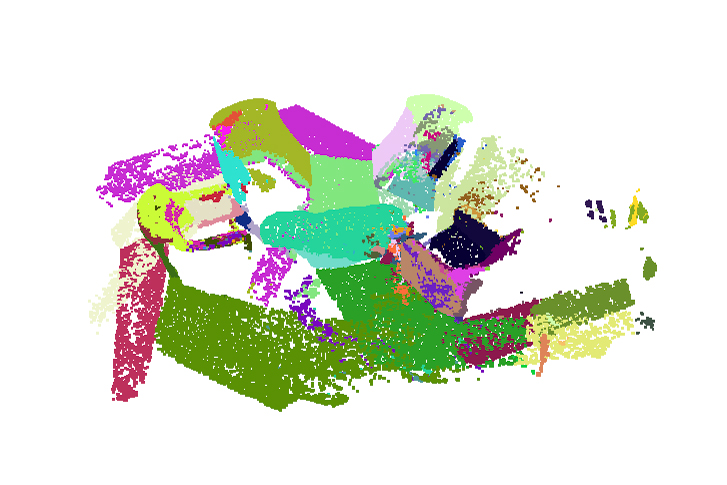}}
        \subfloat[Sparse Labels]{\includegraphics[width=.2\linewidth]{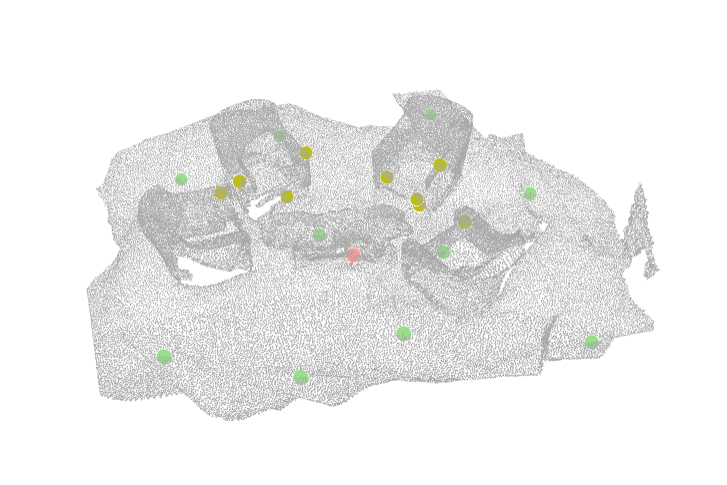}}
        \subfloat[Expanded Labels]{\includegraphics[width=.2\linewidth]{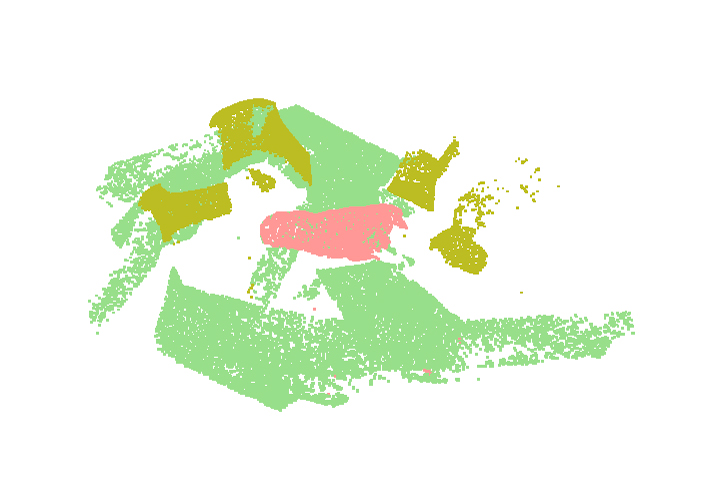}}
        \subfloat[Ground truth]{\includegraphics[width=.2\linewidth]{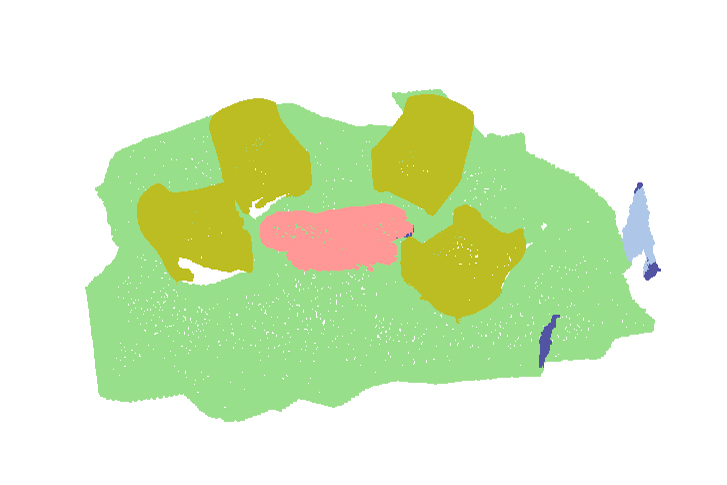}} 

	\caption{Visualization of 3D labels from ScanNetV2 dataset before and after label initialization, with 20 labeled points per scene. The 3D masks projected from the 2D masks accurately represent the position and shape of the 3D objects. The expanded labels are generally accurate compared to the ground truth but do not cover all the masked regions.}
        \label{points}
\end{figure*}

\textbf{Number and Accuracy of Expanded Annotations.} Table \ref{scannet_mask} presents the average number and accuracy of labels within a scene on the ScanNetV2 dataset and the S3DIS dataset, both before and after the process of propagating the initial limited annotations onto the back-projected 3D masks. This demonstrates the efficacy of leveraging 2D images to significantly augment label coverage while maintaining a relatively high level of accuracy.

\begin{table}[htbp]
    \centering
    \caption{The average number and accuracy of labels from ScanNetV2 and S3DIS training set, before and after label initialization. For the ScanNetV2 dataset, the baseline is 20 annotations per scene, but the average is 15 due to some randomly retained points initially lacking ground truth. We also compare the propagation of limited annotations to 3D masks generated by SAM3D \cite{yang_sam3d_2023}. For the S3DIS dataset, we adopt the 'OTOC' setting, randomly retaining 1 annotation per instance, resulting in an average of 36.7 annotations per scene.}
    \begin{tabular}{c|ccc}
        \toprule
        Dataset & Scheme & Number & Accuracy \\
        \midrule
        \multirow{4}{*}{ScanNetV2} & Fully Supervised Annotations & 145170.8 & 100\% \\
                                   & Limited Annotations & 15.0 & 100\% \\
                                   & Expanded Labels (Ours) & 5565.5 & 93.0\% \\
                                   & Expanded Labels (SAM3D) & 62224.5 & 81.8\% \\
        \midrule
        \multirow{3}{*}{S3DIS}    & Fully Supervised Annotations& 955036.4 &100\% \\
                                   & `OTOC' Annotations & 36.7 & 100\% \\
                                   & Expanded Labels (Ours) & 115732.7 & 71.6\% \\
        \bottomrule
    \end{tabular}
    
    \label{scannet_mask}
\end{table}

\begin{figure*}[htbp]
	\centering
        \captionsetup[subfigure]{labelformat=empty}

	\subfloat{\includegraphics[width=.18\linewidth]{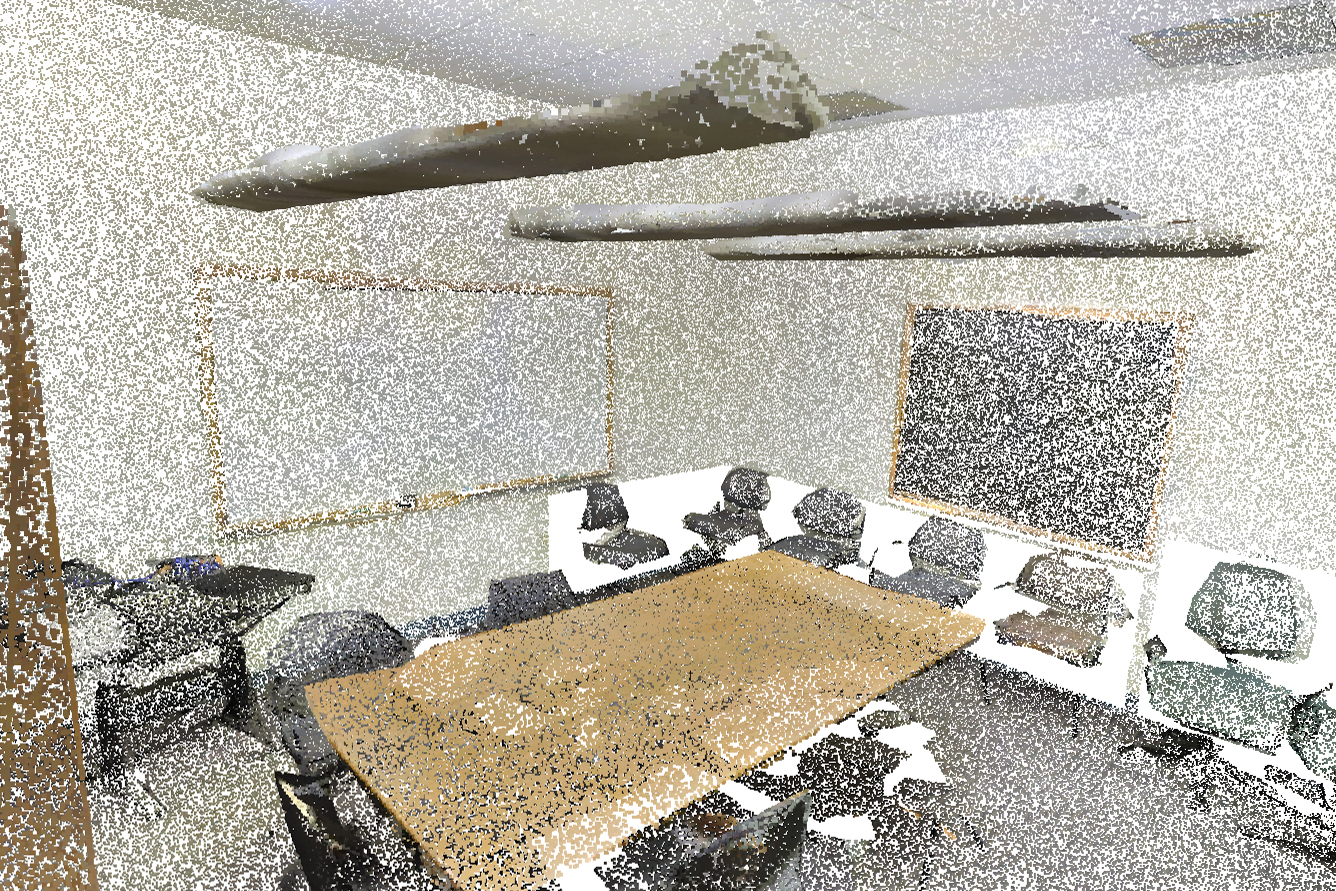}}
    \hspace{1mm}
	\subfloat{\includegraphics[width=.18\linewidth]{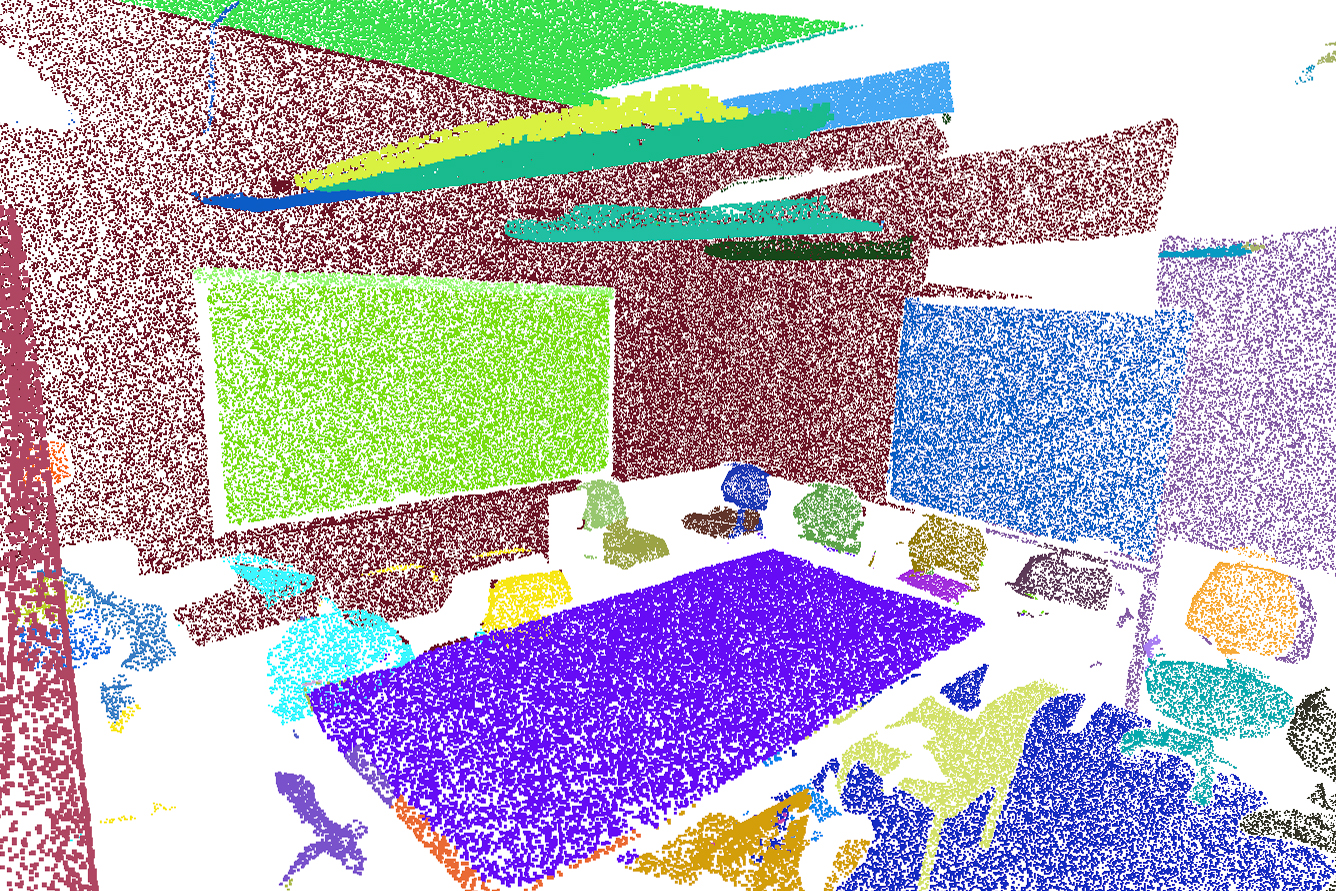}}
    \hspace{1mm}
        \subfloat{\includegraphics[width=.18\linewidth]{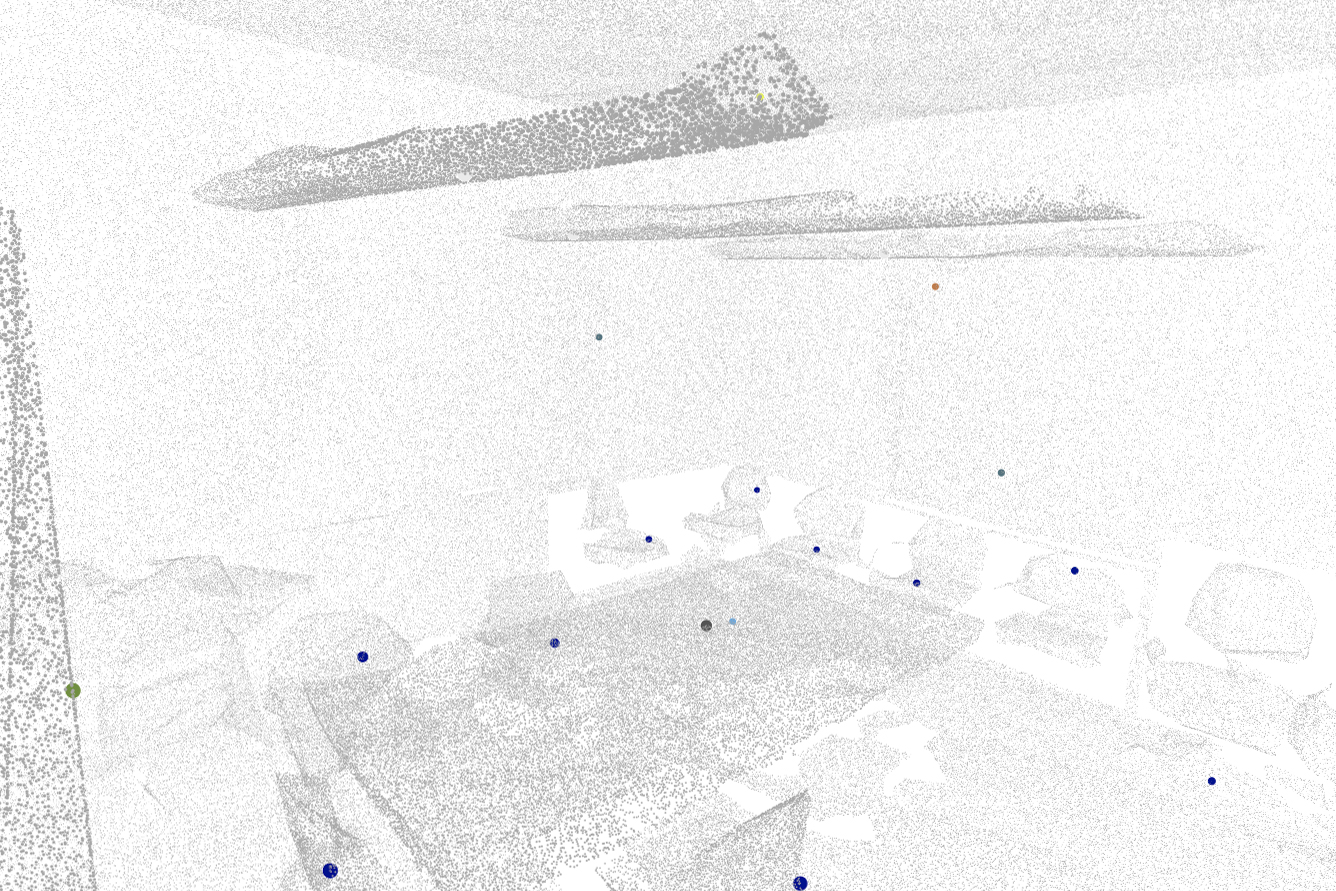}}
        \hspace{1mm}
        \subfloat{\includegraphics[width=.18\linewidth]{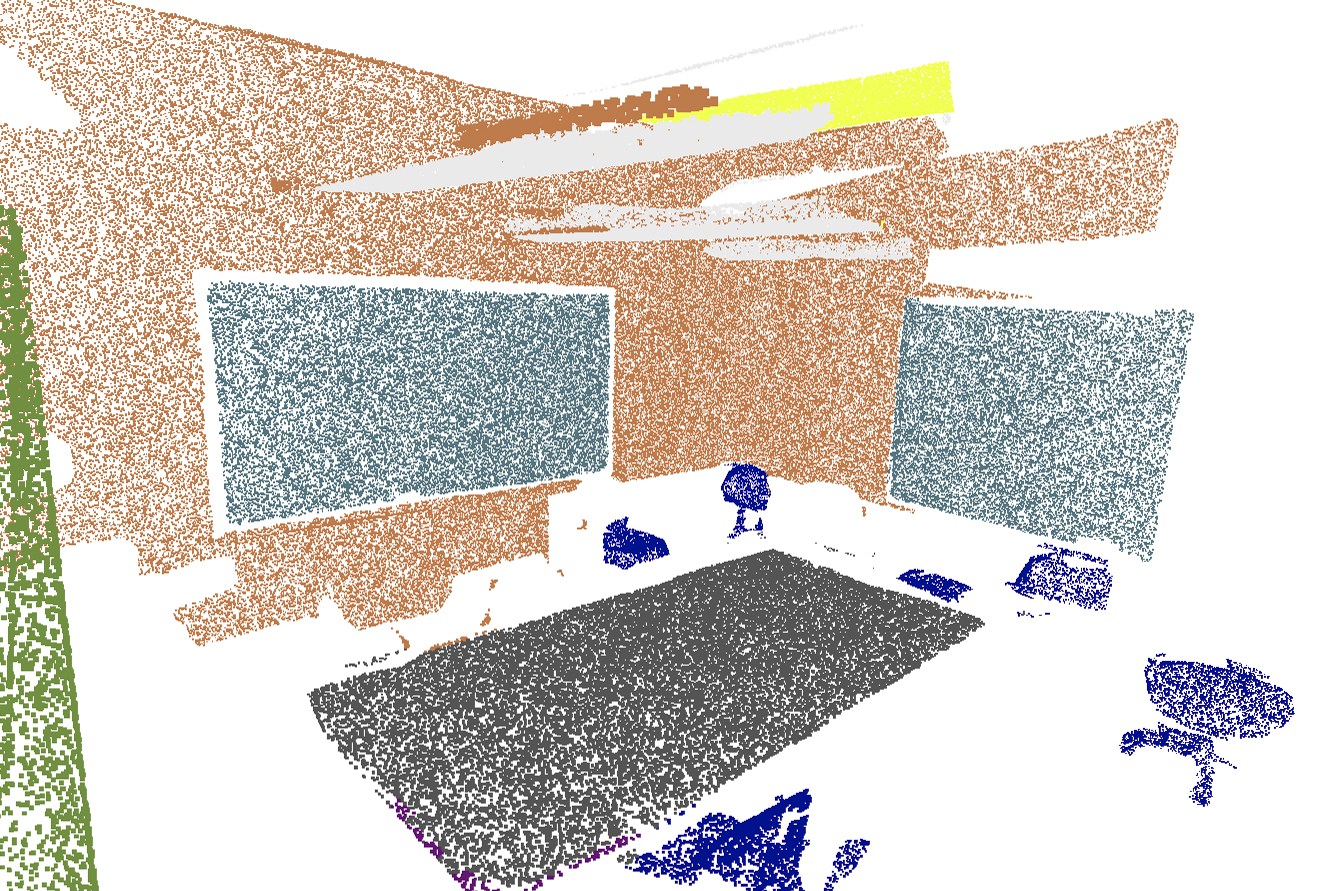}}
        \hspace{1mm}
        \subfloat{\includegraphics[width=.18\linewidth]{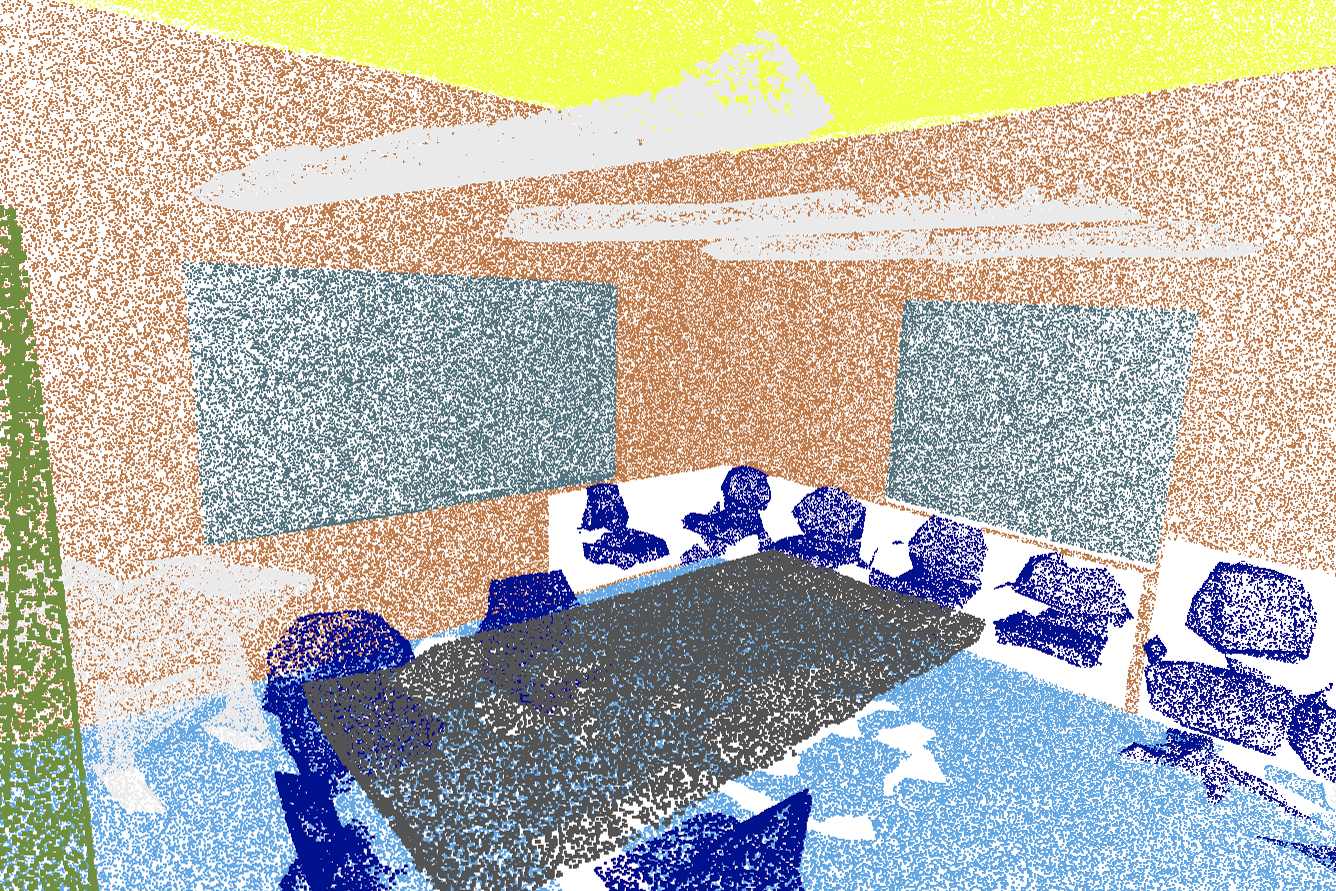}} \\
        \subfloat{\includegraphics[width=.18\linewidth]{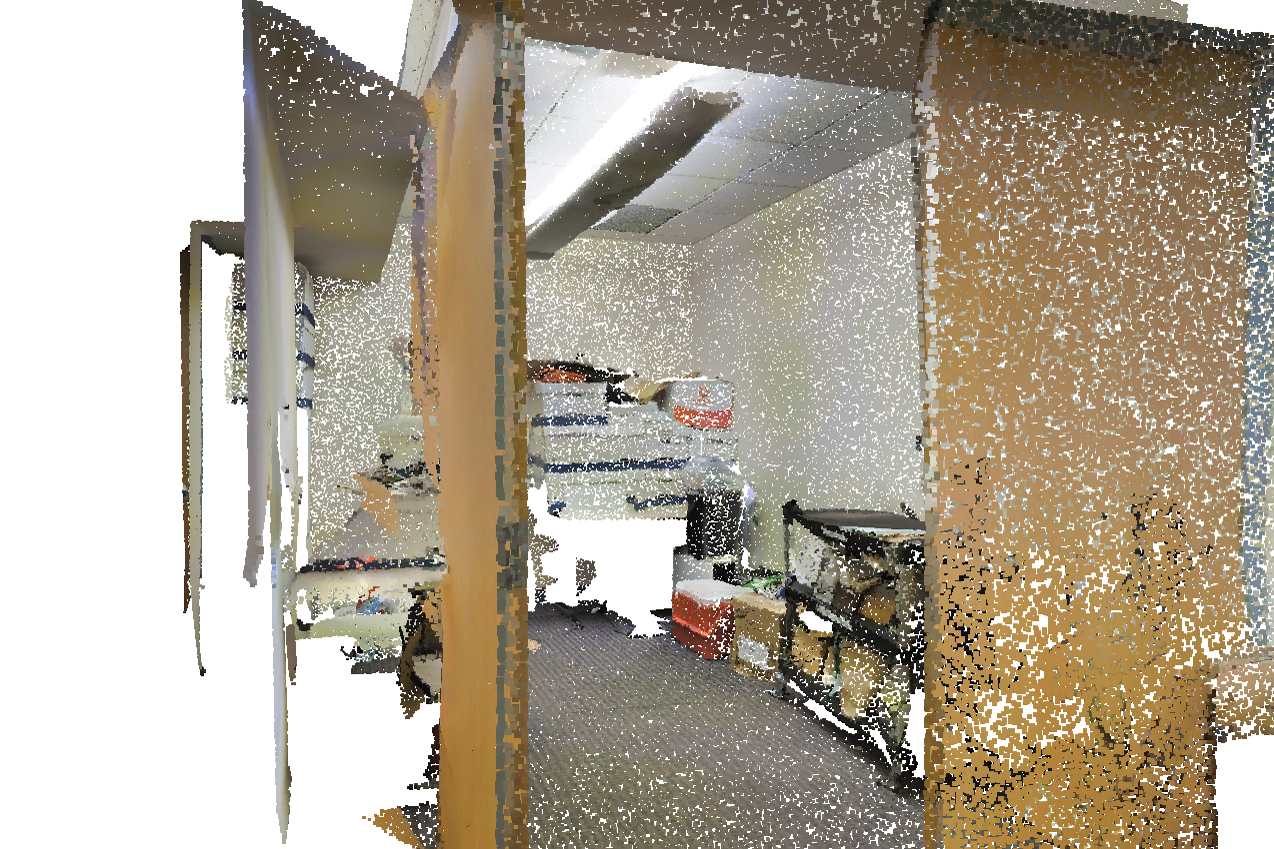}}
        \hspace{1mm}
	\subfloat{\includegraphics[width=.18\linewidth]{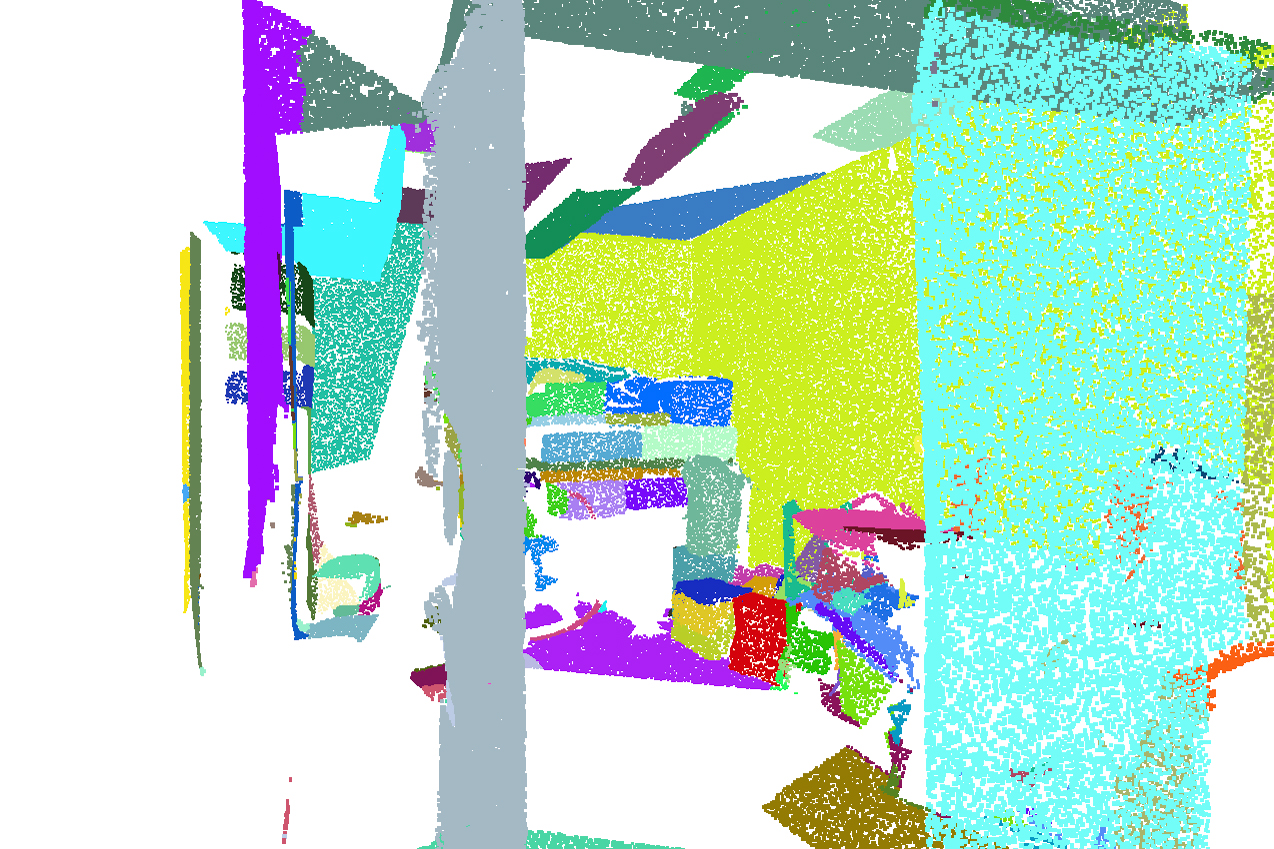}}
    \hspace{1mm}
        \subfloat{\includegraphics[width=.18\linewidth]{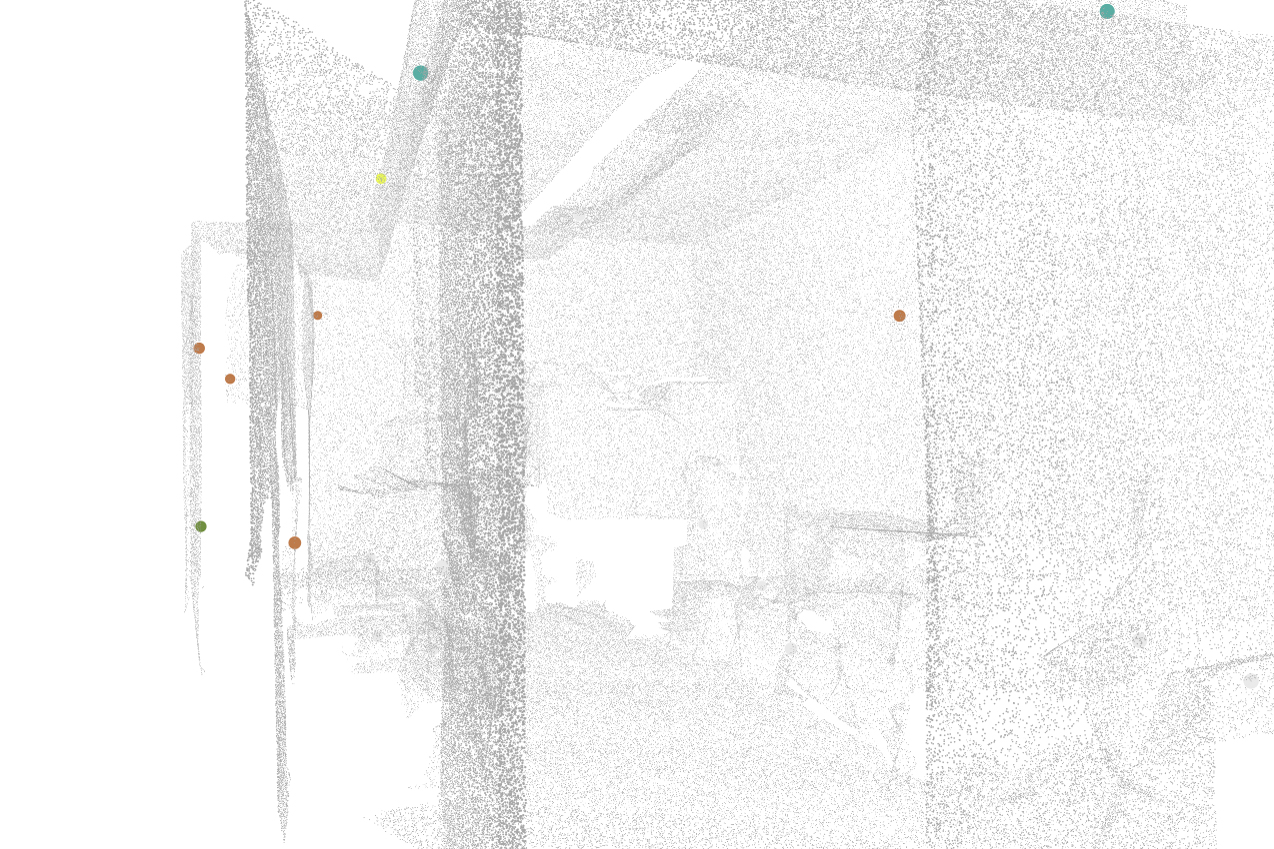}}
        \hspace{1mm}
        \subfloat{\includegraphics[width=.18\linewidth]{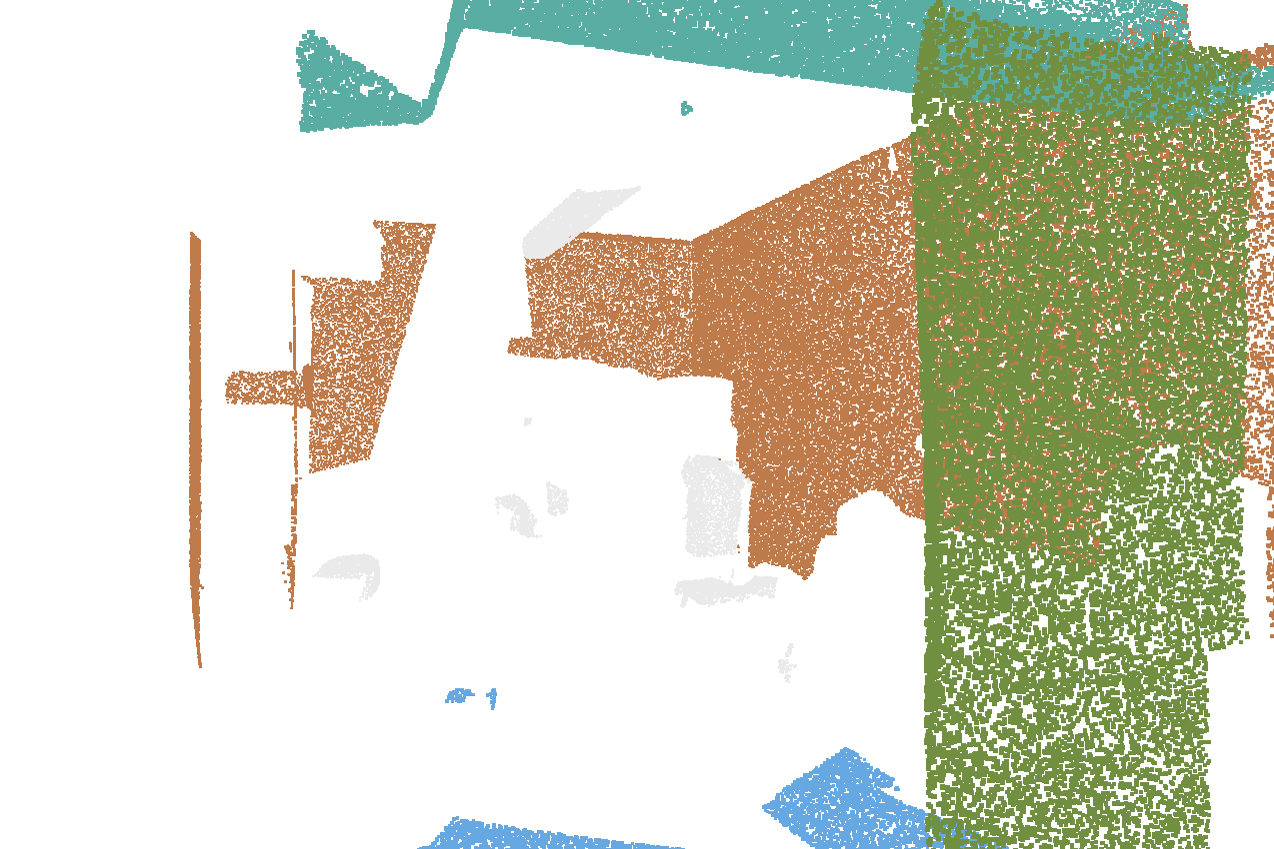}}
        \hspace{1mm}
        \subfloat{\includegraphics[width=.18\linewidth]{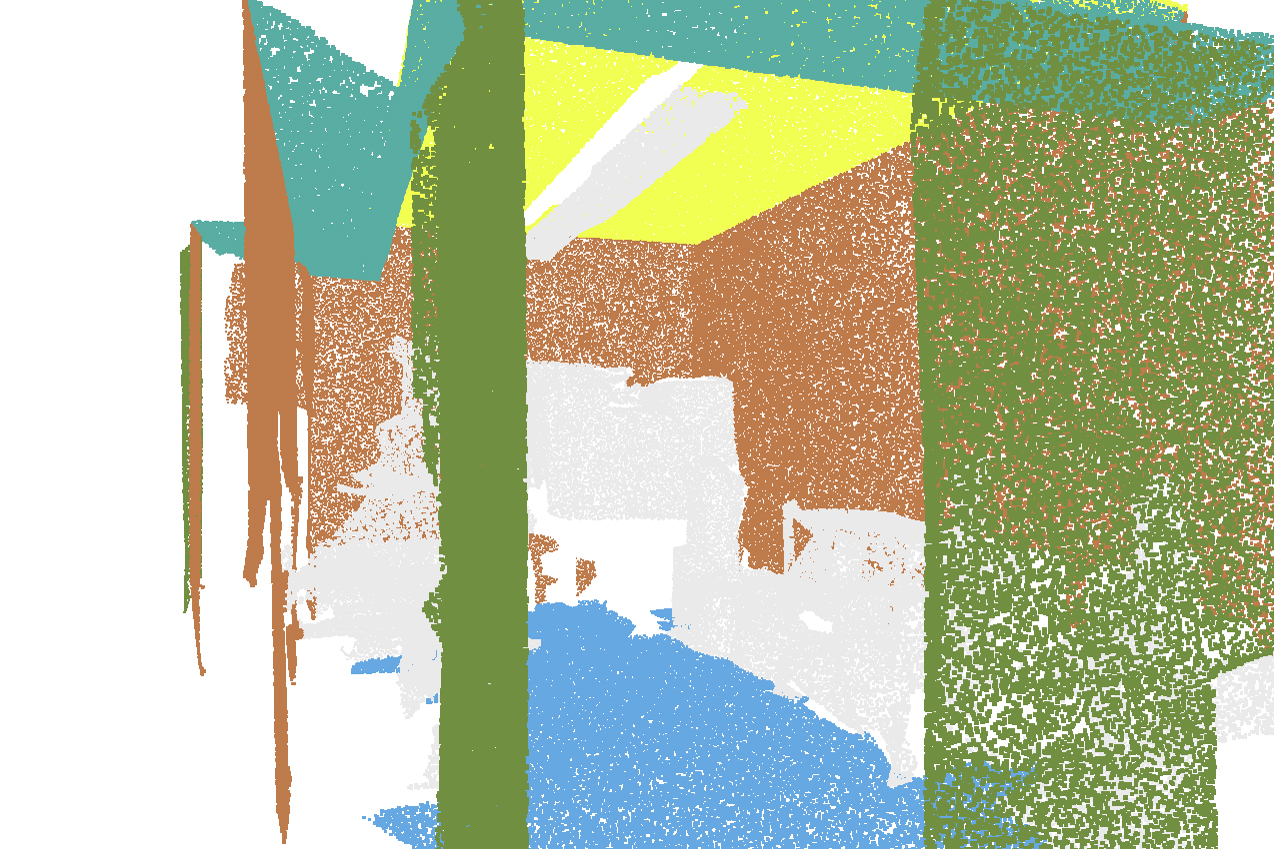}}  \\
        \subfloat[Point Clouds]{\includegraphics[width=.18\linewidth]{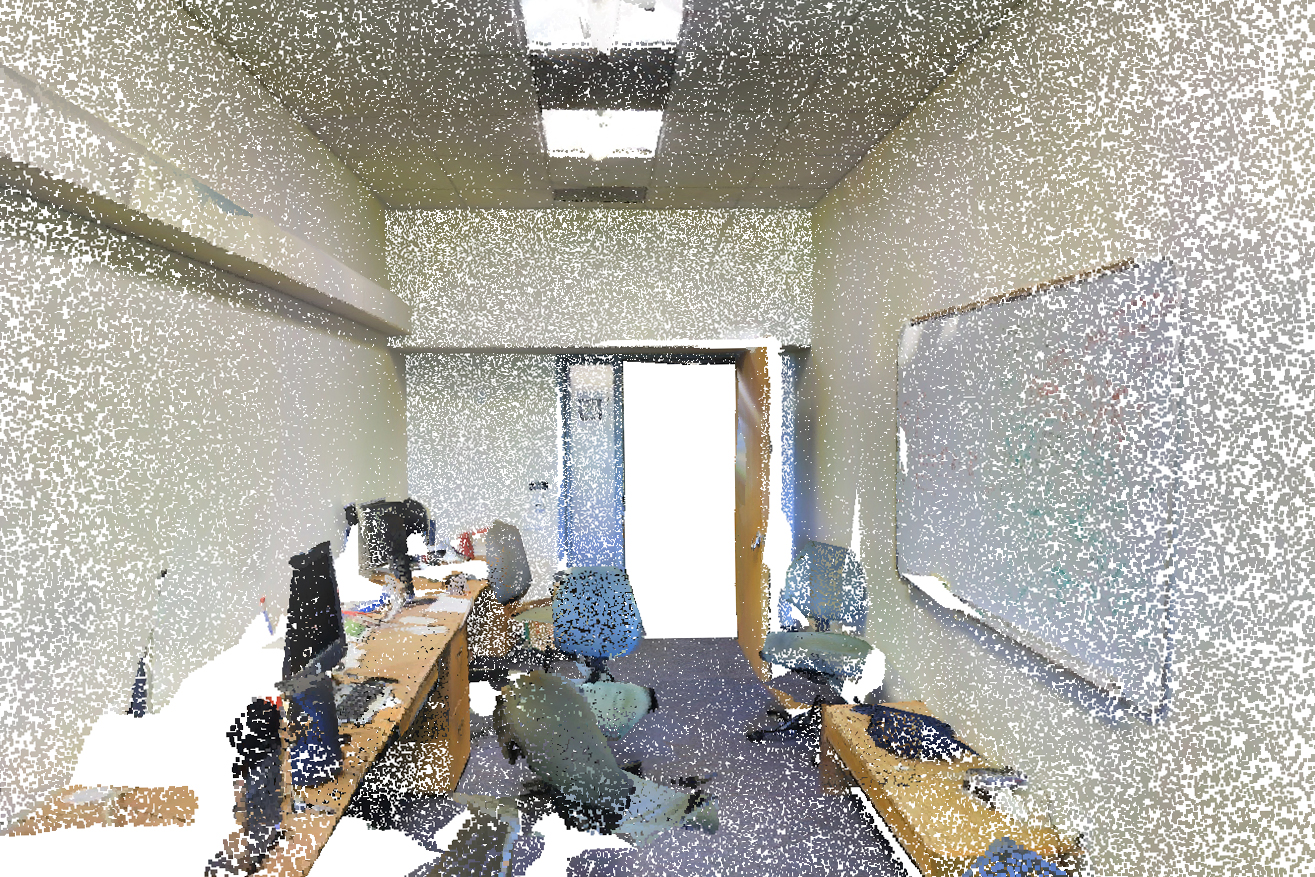}}
        \hspace{1mm}
	\subfloat[3D Masks]{\includegraphics[width=.18\linewidth]{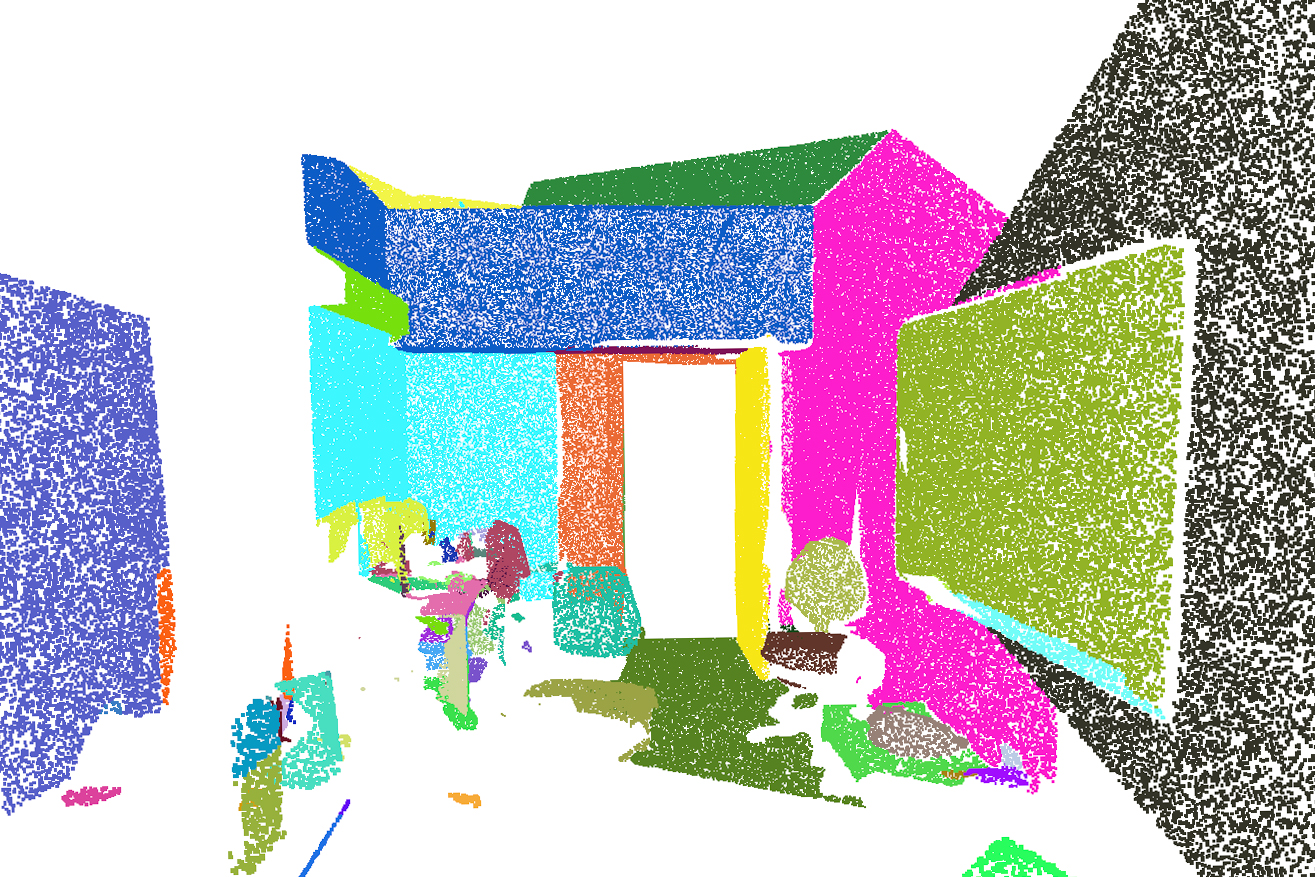}}
        \hspace{1mm}
        \subfloat[Sparse Labels]{\includegraphics[width=.18\linewidth]{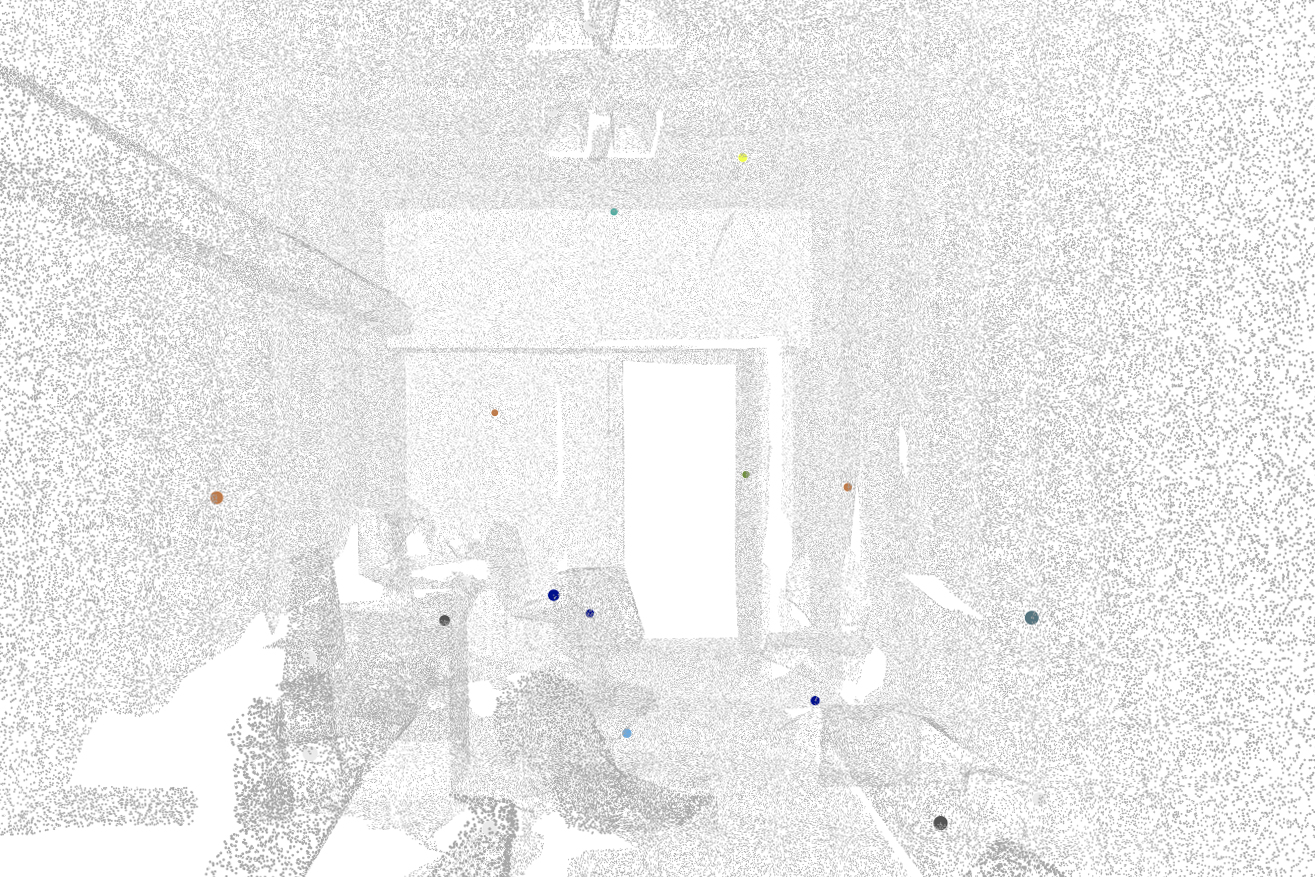}}
        \hspace{1mm}
        \subfloat[Expanded Labels]{\includegraphics[width=.18\linewidth]{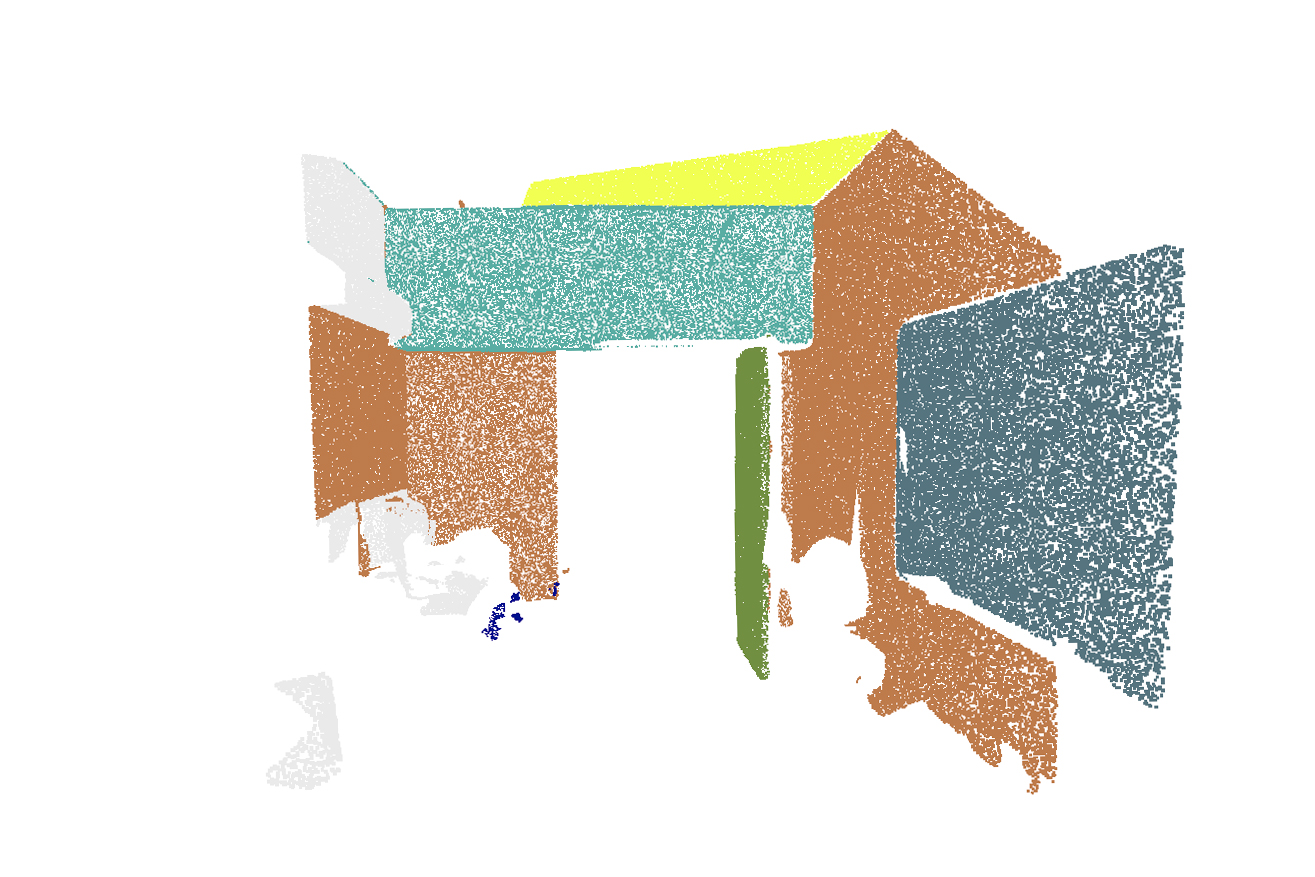}}
        \hspace{1mm}
        \subfloat[Ground truth]{\includegraphics[width=.18\linewidth]{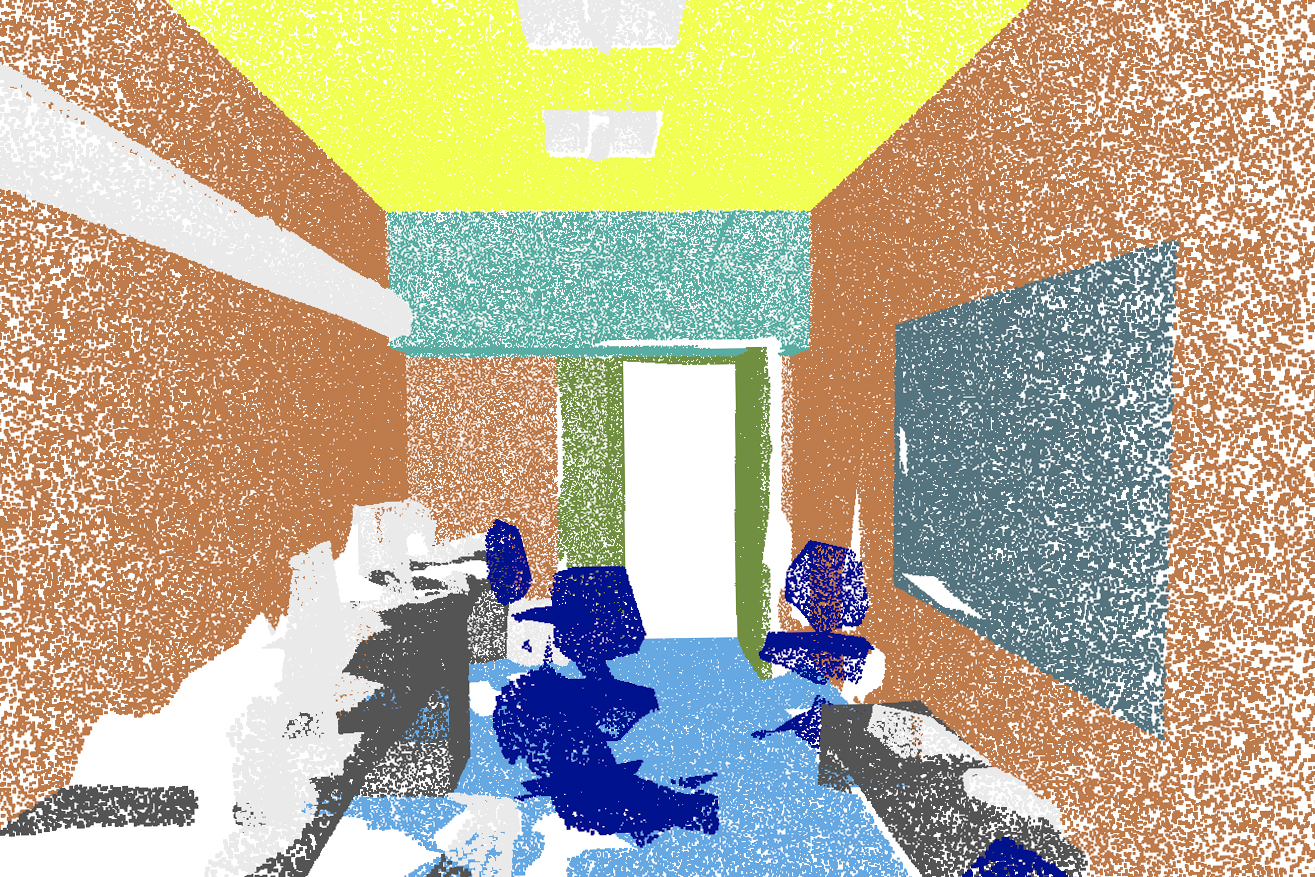}}  \\
	\caption{Visualization of 3D labels from S3DIS dataset before and after label initialization with `OTOC' annotation scheme.}
        \label{points1}
\end{figure*}

\textbf{Visualization of Expanded Labels after Label Initialization.} Some examples of expanded labels obtained by label initialization stage are displayed in Figure \ref{points} and Figure \ref{points1} for ScanNetV2 and S3DIS training set, showing the original point clouds, 3D segmentation masks, initial sparse annotations, expanded labels onto the masks after label initialization, and the ground truth. The expanded labels are mostly accurate compared to the ground truth, but they still cluster around the initial sparse labels. Therefore, in subsequent steps, we select reliable pseudo labels and propagate them onto the masks to fully utilize these masks.

\begin{figure}[htbp]
	\centering
        \captionsetup[subfigure]{labelformat=empty}
        \subfloat[3D masks]{\includegraphics[width=.33\linewidth]{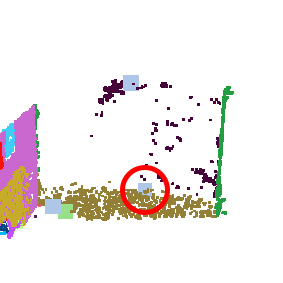}}
	\subfloat[Expanded Labels]{\includegraphics[width=.33\linewidth]{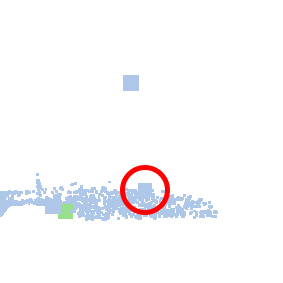}}
        \subfloat[Ground truth]{\includegraphics[width=.33\linewidth]{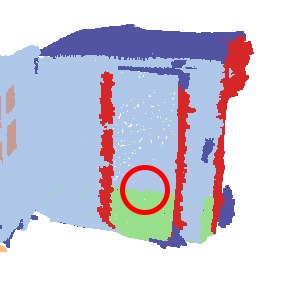}} 
        \caption{A failure case from the ScanNetV2 dataset after label initialization. The circled point is one of the initial limited annotations, labeled as the wall \textcolor[rgb]{0.682, 0.780, 0.910}{\rule{1ex}{1ex}}. However, being located on the boundary between the floor and wall masks, it incorrectly propagates to the floor \textcolor[rgb]{0.596, 0.875, 0.541}{\rule{1ex}{1ex}}.}
        \label{f} 
\end{figure}

Figure \ref{f} illustrates a failure case in label initialization, where a point falling on the boundary is chosen as one of the limited annotations. This label is propagated to another masked region, filling it and introducing noise into the expanded labels.

\begin{figure}[htbp]
	\centering
        \captionsetup[subfigure]{labelformat=empty}
        \subfloat[20 points]{\includegraphics[width=.33\linewidth]{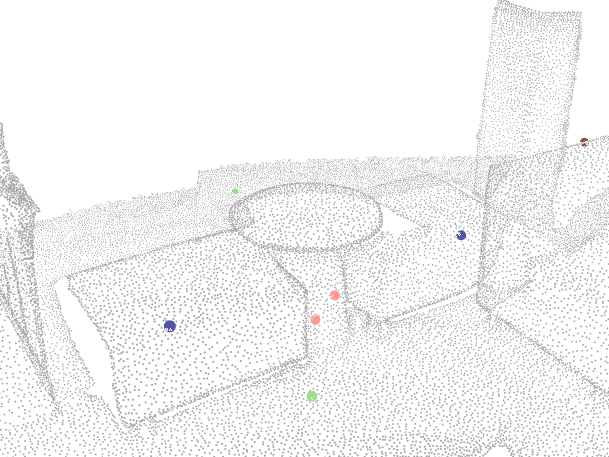}}
        \subfloat[3D masks (SAM3D)]{\includegraphics[width=.33\linewidth]{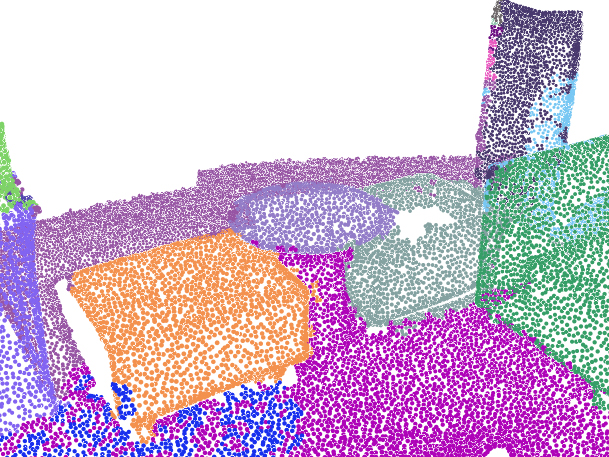}}
        \subfloat[Expanded Labels (SAM3D)]{\includegraphics[width=.33\linewidth]{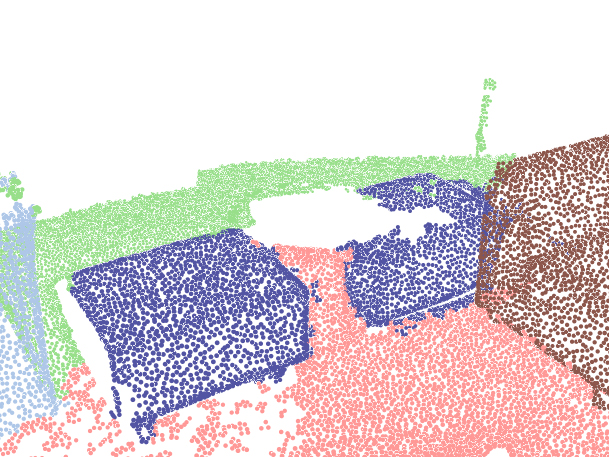}} \\
        \subfloat[Ground truth]{\includegraphics[width=.33\linewidth]{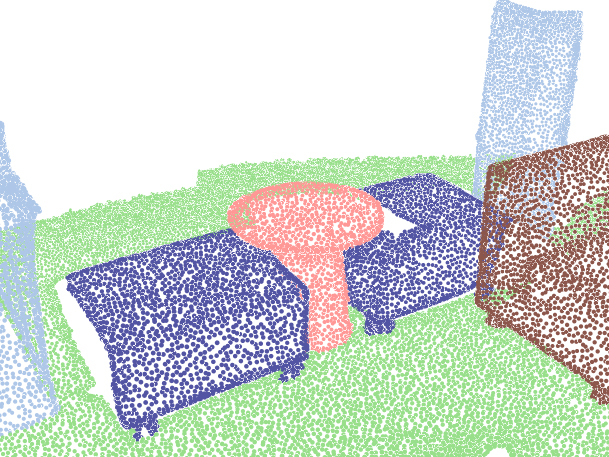}} 
         \subfloat[3D masks (Ours)]{\includegraphics[width=.33\linewidth]{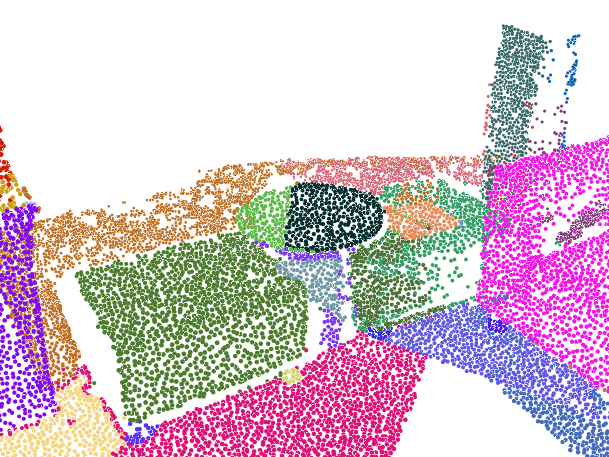}}
        \subfloat[Expanded Labels (Ours)]{\includegraphics[width=.33\linewidth]{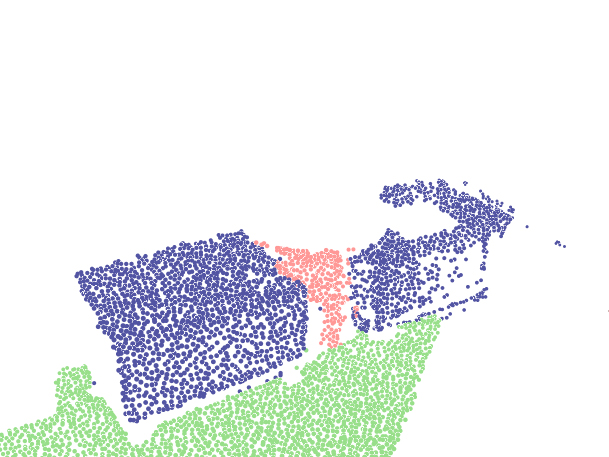}}
        \caption{A failure case occurred when using SAM3D for label initialization, where two masks were incorrectly merged, causing both the floor \textcolor[rgb]{0.596, 0.875, 0.541}{\rule{1ex}{1ex}} and table \textcolor[rgb]{1, 0.596, 0.588}{\rule{1ex}{1ex}} to be treated as table \textcolor[rgb]{1, 0.596, 0.588}{\rule{1ex}{1ex}} when the initial sparse annotations were propagated onto the mask.}
        \label{sam3d} 
\end{figure}

\textbf{Utilizing 3D Masks from SAM3D.} We also explored using segmentation masks generated by SAM3D \cite{yang_sam3d_2023} for 2D images corresponding to each 3D point cloud, as a replacement for our generated masks, and incorporated them into our model's training. SAM3D, by leveraging all RGB images in each scene, produces more complete masks but is more time-consuming. However, it sometimes merges masks of different objects into a single mask, introducing additional noise when limited annotations are propagated onto the masks, as shown in Table \ref{scannet_mask} and Figure \ref{sam3d}. Using masks generated by SAM3D, we trained the model on the ScanNetV2 training set with Point Transformer V3 as the backbone and compared it with our methods on the validation set in Table \ref{scannet_sam3d}.

\begin{table}[h!]
    \centering
    \caption{Comparison on ScanNetV2 validation set by training models using 3D masks generated by our methods and SAM3D.}
    \begin{tabular}{ccc}
        \toprule
         Method & Supervision  & mIoU(\%) \\
         \midrule
         Our Baseline (PTv3) & 20 points & 60.6 \\
        \textbf{SAM3D} & 20 points & \textbf{70.2} \\
        \textbf{Ours} & 20 points & \textbf{71.3} \\
        Our Upper Bound (PTv3) & 100\%  & 77.5 \\
        \bottomrule
    \end{tabular}
    
    \label{scannet_sam3d}
\end{table}

\subsubsection{Noise-Robust Loss on the Expanded Labels}

The expanded sparse annotations can be utilized directly by incorporating an additional loss term alongside the Cross-Entropy loss for original sparse labels utilized in the baseline method, without the need for additional techniques, i.e. $\mathbf{L = \lambda_{seg}\mathbf{L_{seg}}+\lambda_{m}\mathbf{L_m}}$. 

Because reality lacks ground truth, and one point may not adequately represent the entire mask due to noise in projected masks or misalignment among classes, the accuracy of expanded labels is uncertain. Therefore, they are regarded as noisy labels. As a result, noise-robust loss functions \cite{ma_normalized_2020, zhou_asymmetric_2021} are compared on the ScanNetV2 dataset on Point Transformer V3 backbone, as shown in Table \ref{scannet_norm}. Moreover, Table \ref{s3dis_norm} shows the results on the S3DIS dataset. These results demonstrate that despite the noise in the expanded labels, their direct application with an additional loss can enhance performance. This also proves the efficacy of propagating limited annotations onto back-projected masks.

\begin{table}[htbp]
    \centering
    \caption{Results on ScanNetV2 validation set with the noise-robust loss on expanded sparse annotations. In particular, the `NCE' and `RCE’ losses achieved the highest results, so we chose to apply them to all labels expanded to 3D masks.}
    \begin{tabular}{ccccccccc}
        \toprule
        CE & NFL & MAE & RCE & NCE & AGCE & AUE & mIoU (\%) \\
         \midrule
         - &-  &-  &-  &-  &-  &-  & 60.6 \\
         \checkmark &-  &-  &-  &-  &-  &-  & 62.4 \\
         - & \checkmark  & \checkmark  &-  &-  &-  &-  & 60.8 \\
         - & \checkmark  & -  & \checkmark  &-  &-  &-  & 62.8 \\
         - & -  & \checkmark & - & \checkmark  &-  &-  & 60.9 \\
         - & -  & -  & \checkmark & \checkmark  & -  & -  & \textbf{63.0} \\
         - & -  & -  & - & \checkmark  & \checkmark  &-  & 62.3 \\
         - & -  & -  & - & \checkmark  & -  & \checkmark  & 61.4 \\
        \bottomrule
    \end{tabular}
    
    \label{scannet_norm}
\end{table}
\begin{table}[htbp]
    \centering
    \caption{Results on S3DIS testing set with noise-robust loss on expanded sparse annotations.}
    \begin{tabular}{ccccccccc}
        \toprule
        CE & NFL & MAE & RCE & NCE & AGCE & AUE & mIoU (\%) \\
         \midrule
         - &-  &-  &-  &-  &-  &-  & 54.7 \\
         \checkmark &-  &-  &-  &-  &-  &- & 56.3 \\
         - & -  & -  & \checkmark & \checkmark  &-  &-  & \textbf{57.1} \\
        \bottomrule
    \end{tabular}
    \label{s3dis_norm}
\end{table}

\subsubsection{Selection of $\eta$}

As mentioned in Section \ref{init}, for reliable pseudo labels, due to their potential inaccuracy or instability, it is desirable for them to occupy a certain proportion of each mask-represented region, and then expand them to the entire mask. The hyperparameter $\eta$ is introduced to control this proportion. 

\textbf{Comparison on different values of $\eta$.} On the one hand, it is desirable to expand reliable pseudo labels as much as possible; on the other hand, the higher the proportion they occupy on the mask, the more reliable the expanded labels are. Table \ref{eta} demonstrates the effects of training the model with different values of $\eta$. 

\begin{table}[htbp]
    \centering
    \caption{Comparison on different value of $\eta$ on ScanNetV2 validation set.}
    \begin{tabular}{cc}
        \toprule
        $\eta_0$ & mIoU (\%)\\
         \midrule
          0.3 & 70.2 \\
         0.5 & 70.4 \\
         0.7 &  \textbf{71.3}\\
         0.9 &  70.7 \\
        \bottomrule
    \end{tabular}
    
    \label{eta}
\end{table}

\begin{figure}[htbp]
    \centering
    \includegraphics[width=.7\linewidth]{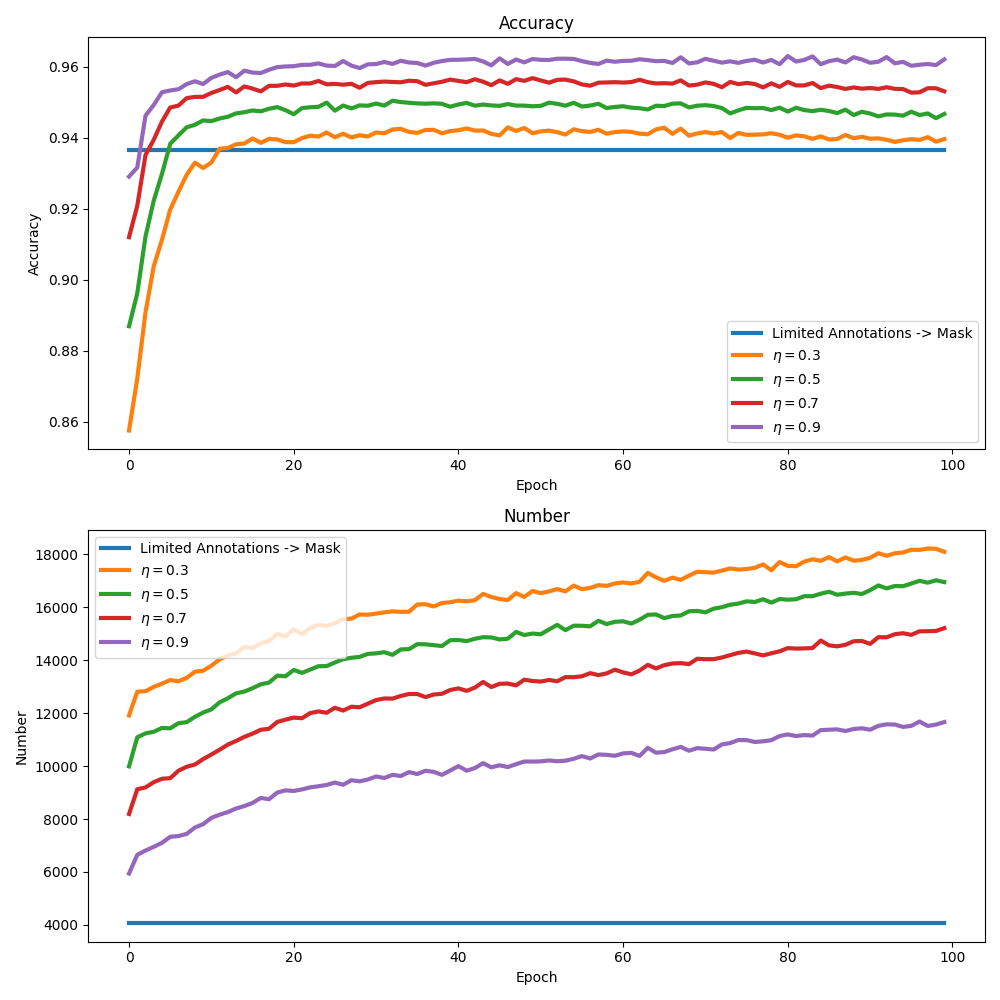}
    \caption{Number and accuracy of $\widetilde{Y}$ with different value of $\eta$ through the training process.}
    \label{pseudo}
\end{figure}

\textbf{Number and Accuracy of Expanded Labels $\widetilde{Y}$.}
Figure \ref{pseudo} shows the number and accuracy of the expanded limited annotations and reliable pseudo labels after propagating them onto the masks, defined as $\widetilde{Y}$. From the figure, it can be observed that the accuracy of $\widetilde{Y}$ tends to stabilize in the later stages, with its quantity increasing with more epochs. 
\subsubsection{Ablations}
In this section, we assess the effectiveness of each loss function and the corresponding computational entities within the total objective function on the ScanNetV2 dataset, with 20 training points in each scene. The results of the ScanNetV2 validation set are shown in Table \ref{ablation}. These results demonstrate the effectiveness of the different losses and their components.

\begin{figure}[h!]
	\centering
        \captionsetup[subfigure]{labelformat=empty}
        \subfloat[20 points]{\includegraphics[width=.4\linewidth]{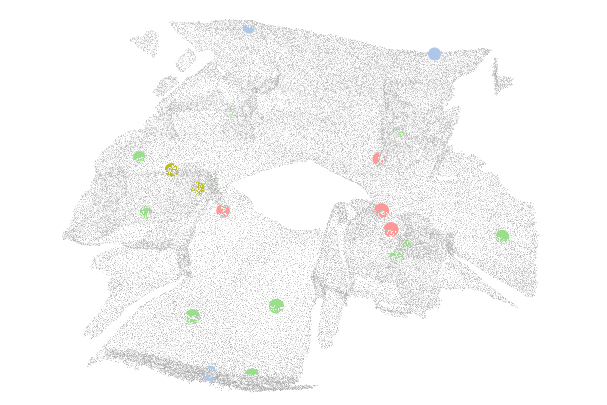}}
        \subfloat[20 points $\rightarrow Mask_{3D}$]{\includegraphics[width=.4\linewidth]{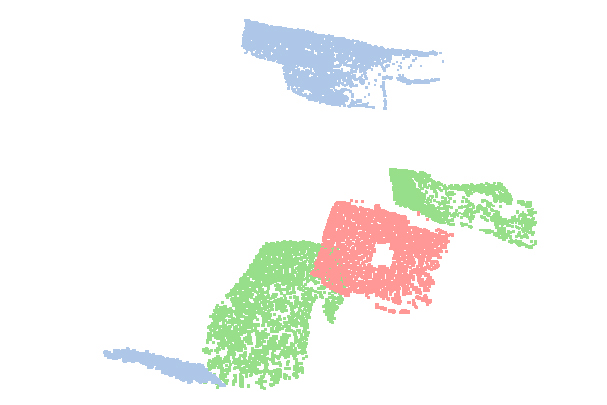}} \\
        \subfloat[\centering 20 points + reliable set $\rightarrow Mask_{3D}$]
        {\includegraphics[width=.4\linewidth]{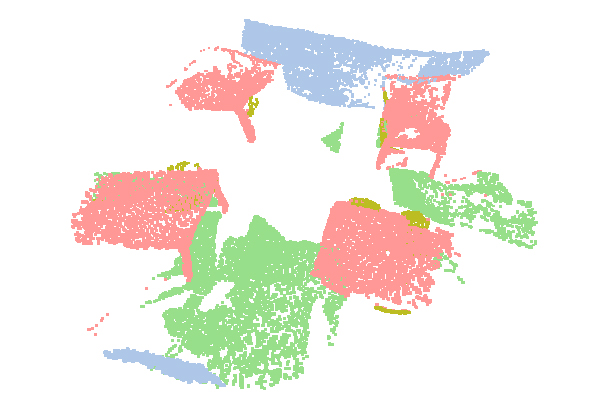}}
        \subfloat[Ground truth]{\includegraphics[width=.4\linewidth]{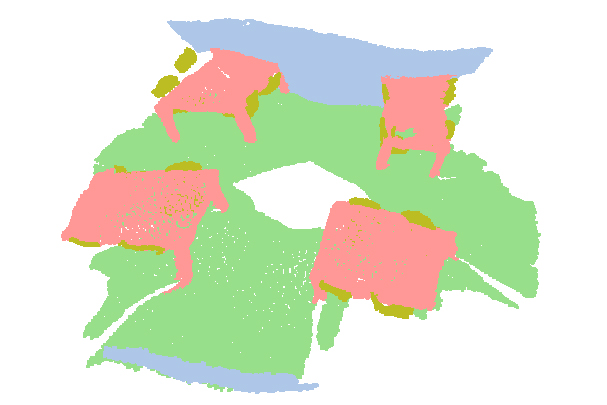}} 
        
        \caption{An example of expanded labels from the ScanNetV2 dataset. The available labels are greatly expanded after propagating initial annotations and reliable pseudo labels to the masks.}
        \label{expand} 
\end{figure}

Corresponding to these computational entities, Figure \ref{expand} shows an example of the expanded labels and ground truth. This demonstrates that propagating the initial sparse annotations to the masks and subsequently adding reliable pseudo labels to the masks can expand the available labels with high accuracy.

\begin{table*}[htbp]
    \centering
    \caption{Ablations on ScanNetV2 validation set.}
    \begin{tabular}{ccccc}
        \toprule
         CE Loss & NCEandRCE &CE Loss & KL Loss &mIoU(\%) \\
         \midrule
         20 points & - & - & - & 60.6\\
         20 points & 20 points $\rightarrow Mask_{3D}$ & - & - & 63.0 \\
         20 points & - & reliable set & ambiguous set & 69.9\\
         20 points & 20 points $\rightarrow Mask_{3D}$  & reliable set & ambiguous set &  70.8 \\
         20 points & reliable set $\rightarrow Mask_{3D}$  & reliable set & ambiguous set& 70.4 \\
         20 points & 20 points + reliable set $\rightarrow Mask_{3D}$  & reliable set & ambiguous set & 71.3\\
        \bottomrule
    \end{tabular}
    
    \label{ablation}
\end{table*}

\subsubsection{Result Visualization}
\begin{figure}[h!]
	\centering
         \captionsetup[subfigure]{labelformat=empty}

        \subfloat{\includegraphics[width=.3\linewidth]{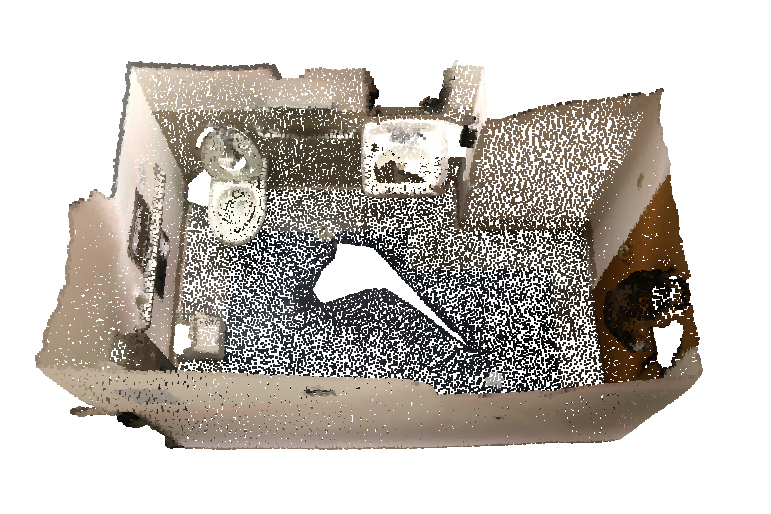}}
	\subfloat{\includegraphics[width=.3\linewidth]{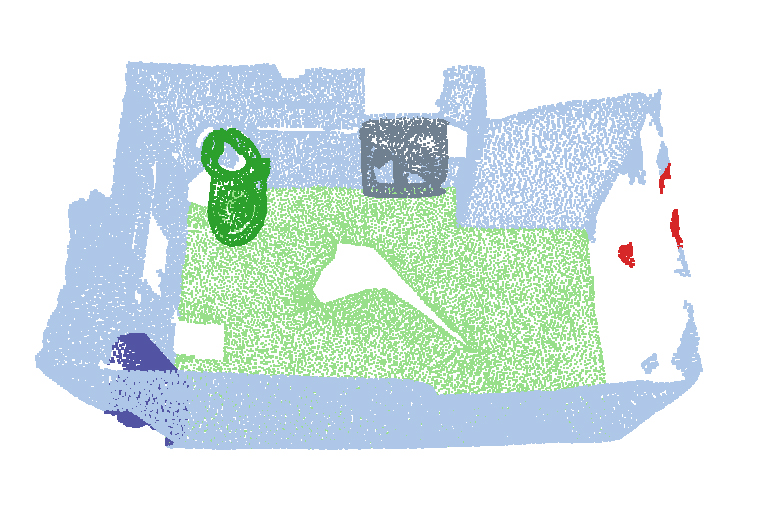}}
        \subfloat{\includegraphics[width=.3\linewidth]{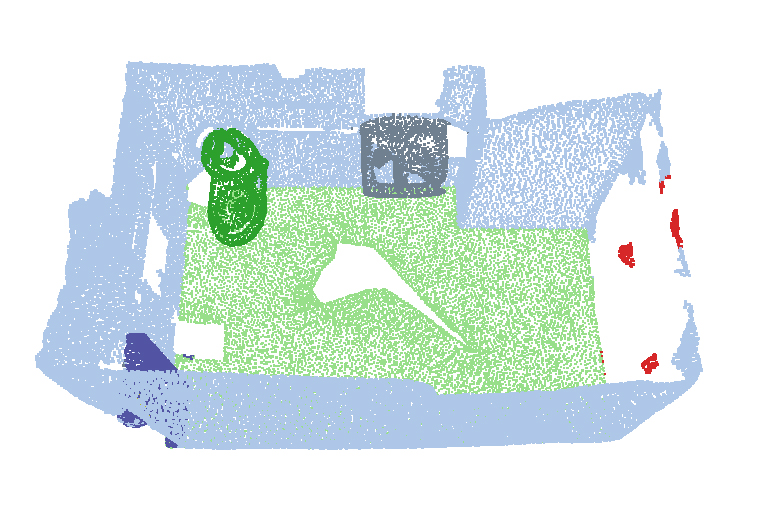}} \\

        \subfloat{\includegraphics[width=.3\linewidth]{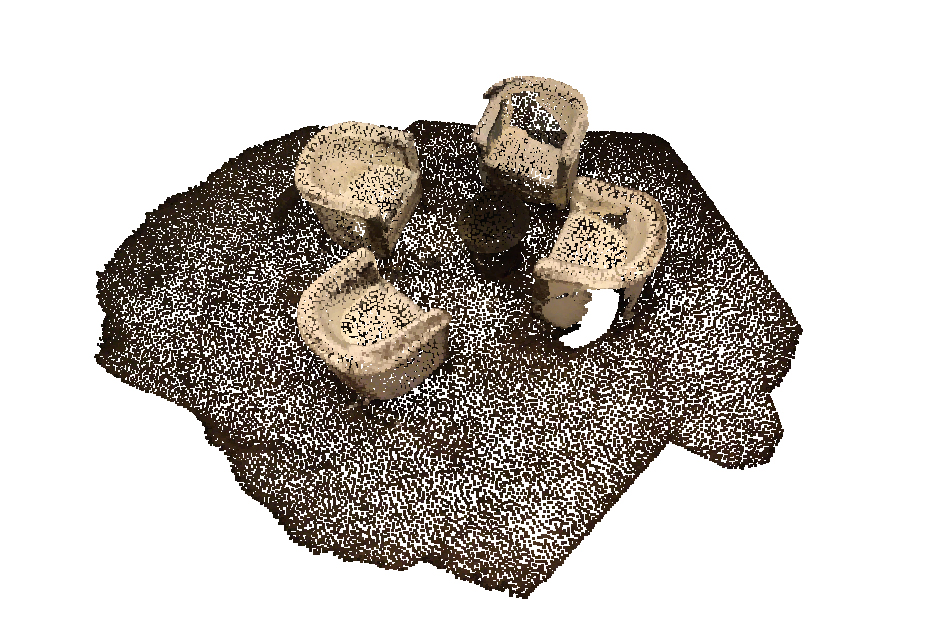}}
	\subfloat{\includegraphics[width=.3\linewidth]{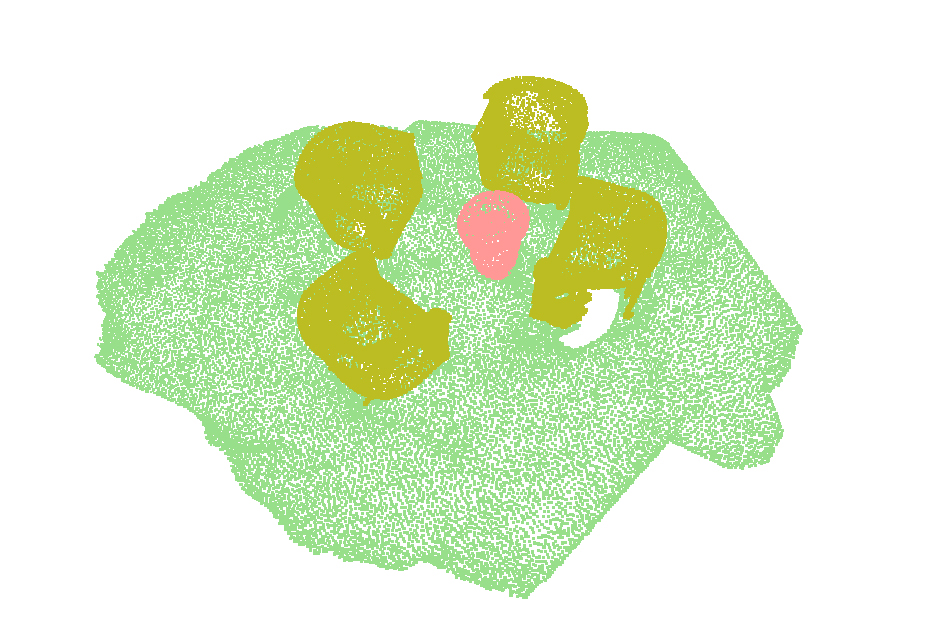}}
        \subfloat{\includegraphics[width=.3\linewidth]{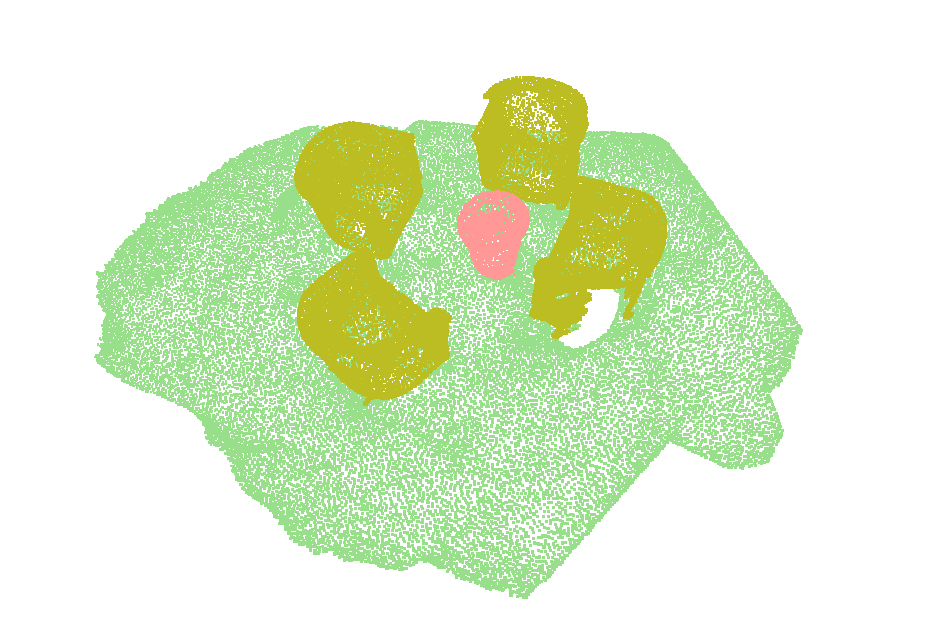}} \\

        \subfloat[Point Clouds]{\includegraphics[width=.3\linewidth]{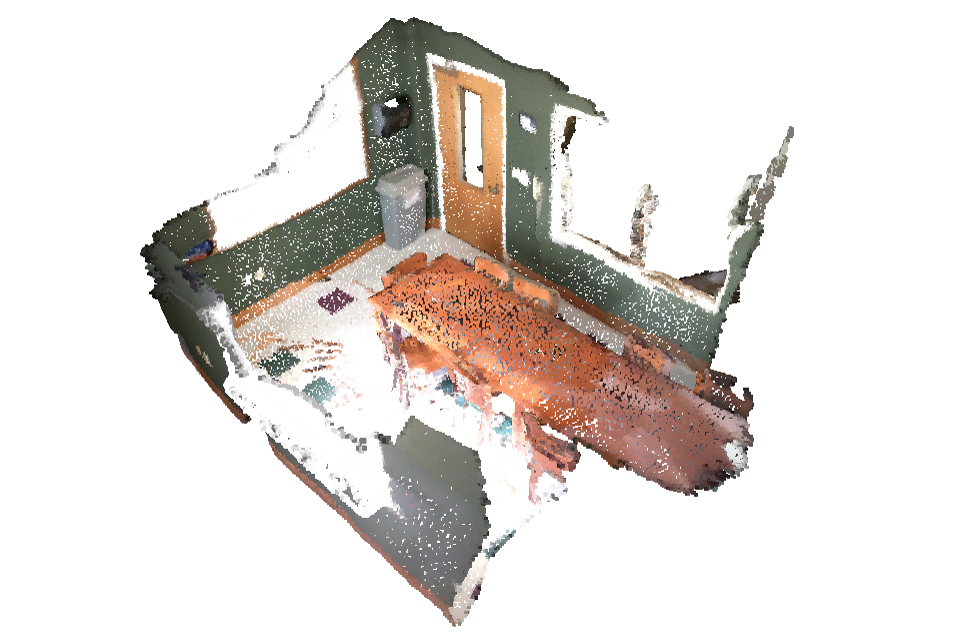}}
	\subfloat[Ground truth]{\includegraphics[width=.3\linewidth]{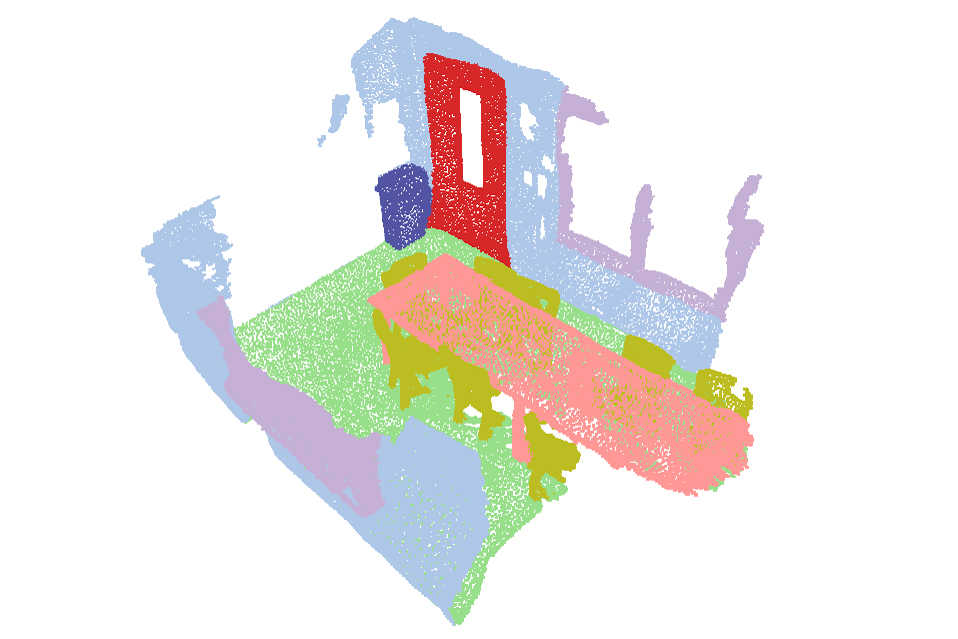}}
        \subfloat[Prediction]{\includegraphics[width=.3\linewidth]{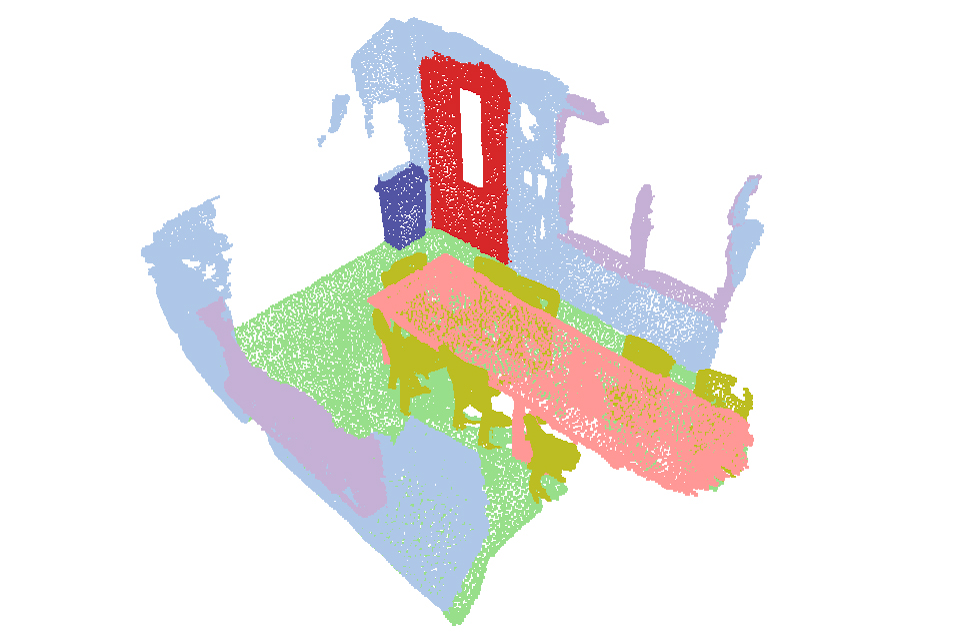}} \\
        
        \caption{Examples of semantic segmentation results on ScanNetV2 validation set. Our model can perform semantic segmentation with high accuracy, but there are still some issues with edges and small objects.}
        \label{result_scannet} 
\end{figure}

\begin{figure}[h!]
	\centering
         \captionsetup[subfigure]{labelformat=empty}
        \subfloat{\includegraphics[width=.3\linewidth]{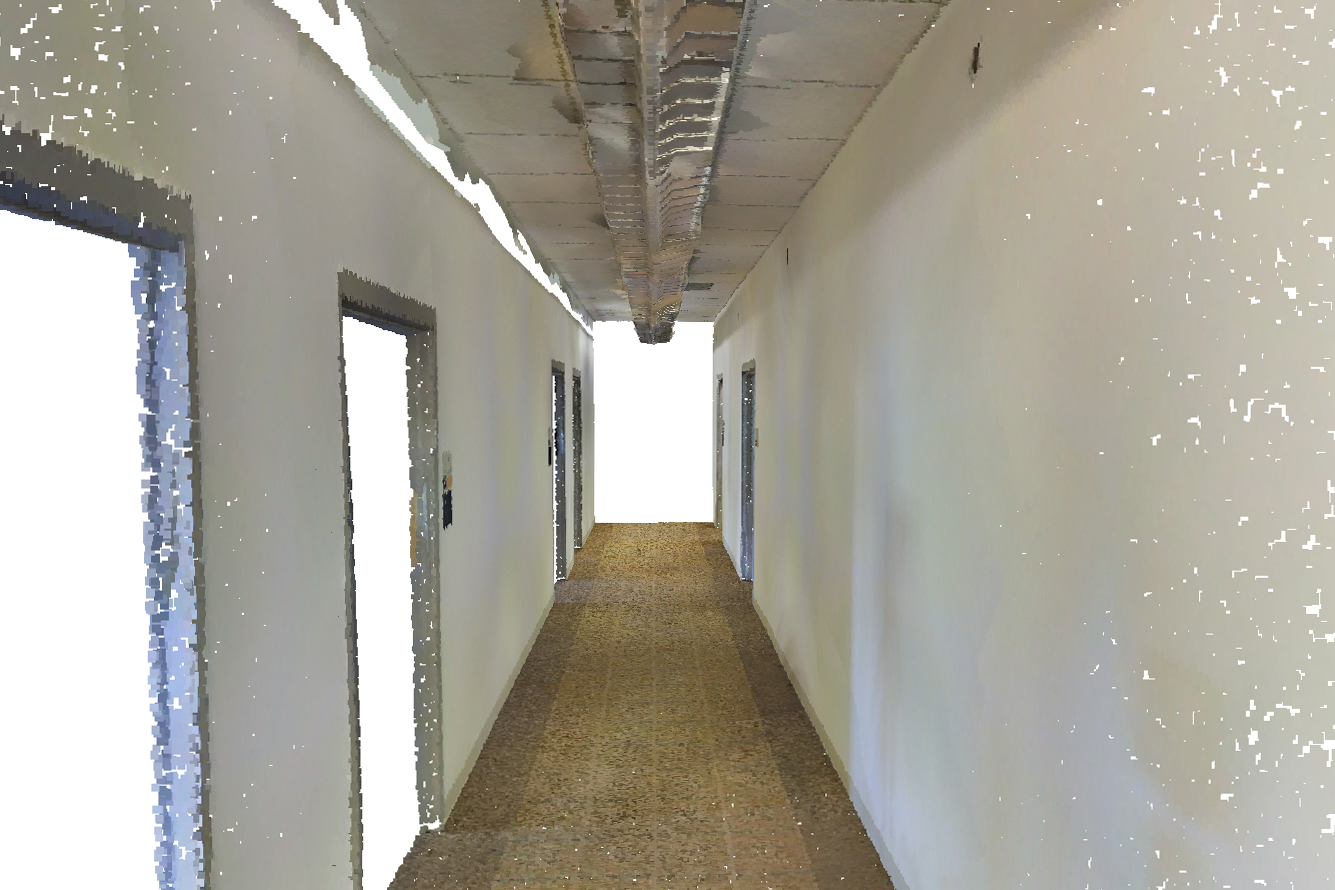}}
	\subfloat{\includegraphics[width=.3\linewidth]{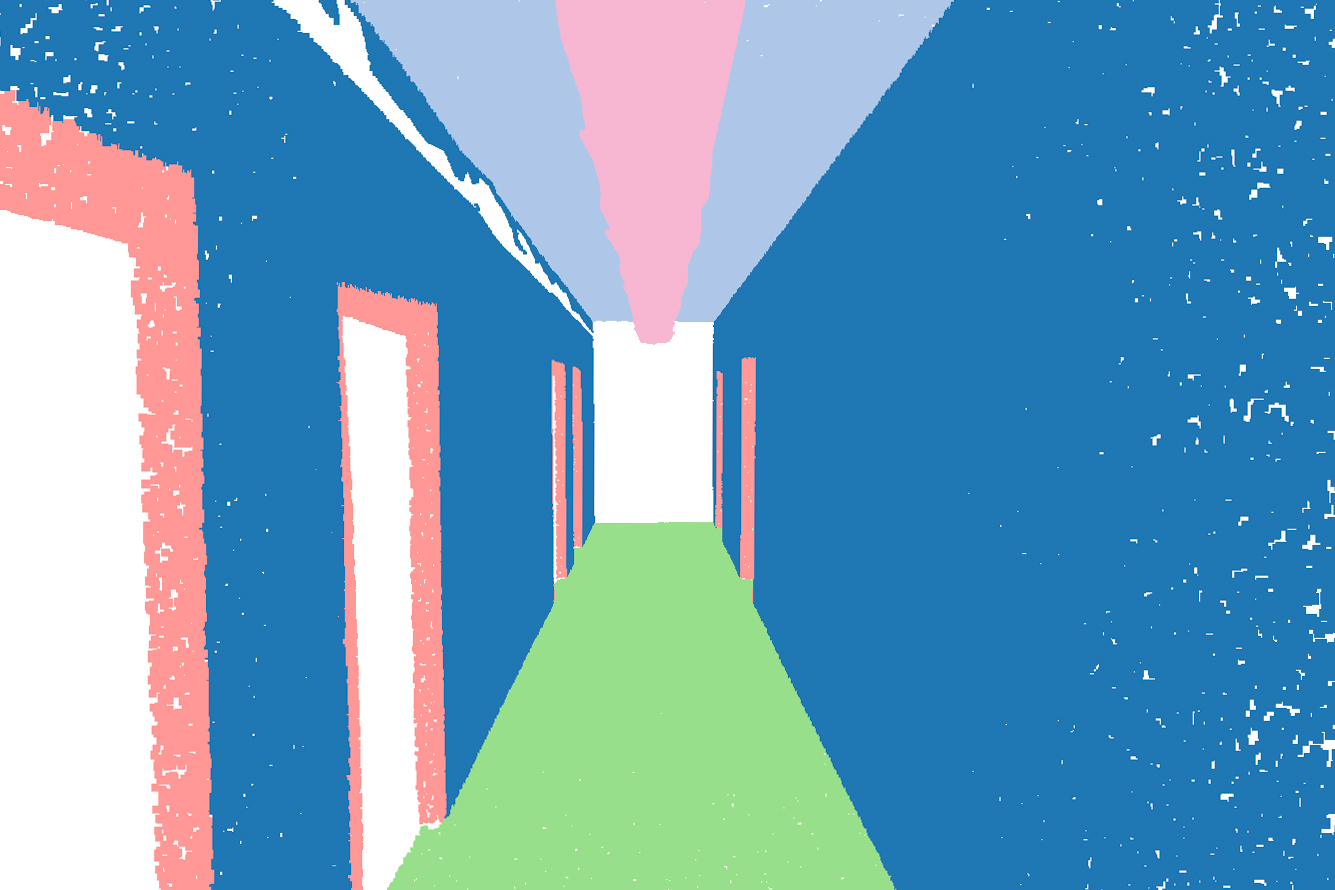}}
        \subfloat{\includegraphics[width=.3\linewidth]{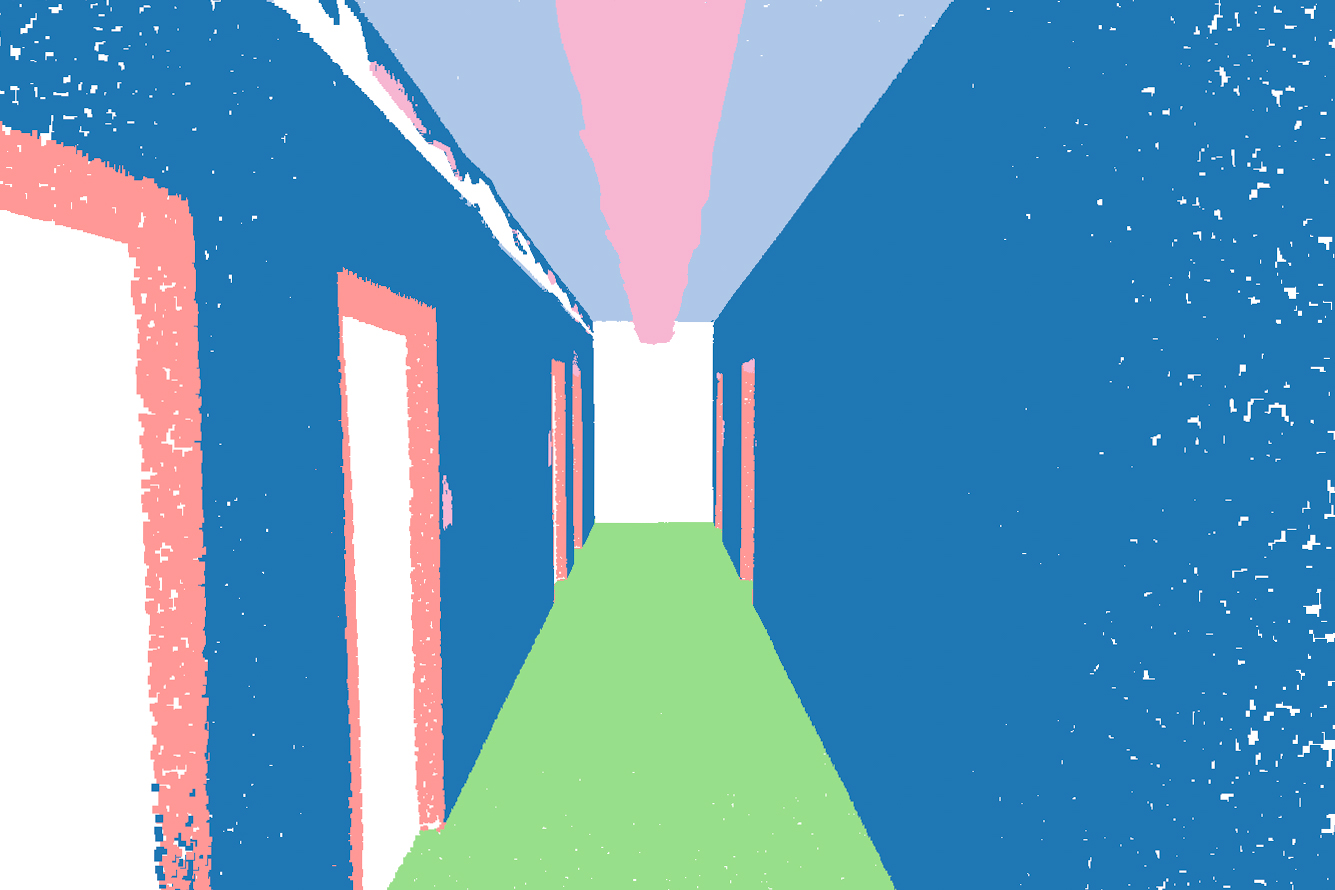}} \\
        \subfloat{\includegraphics[width=.3\linewidth]{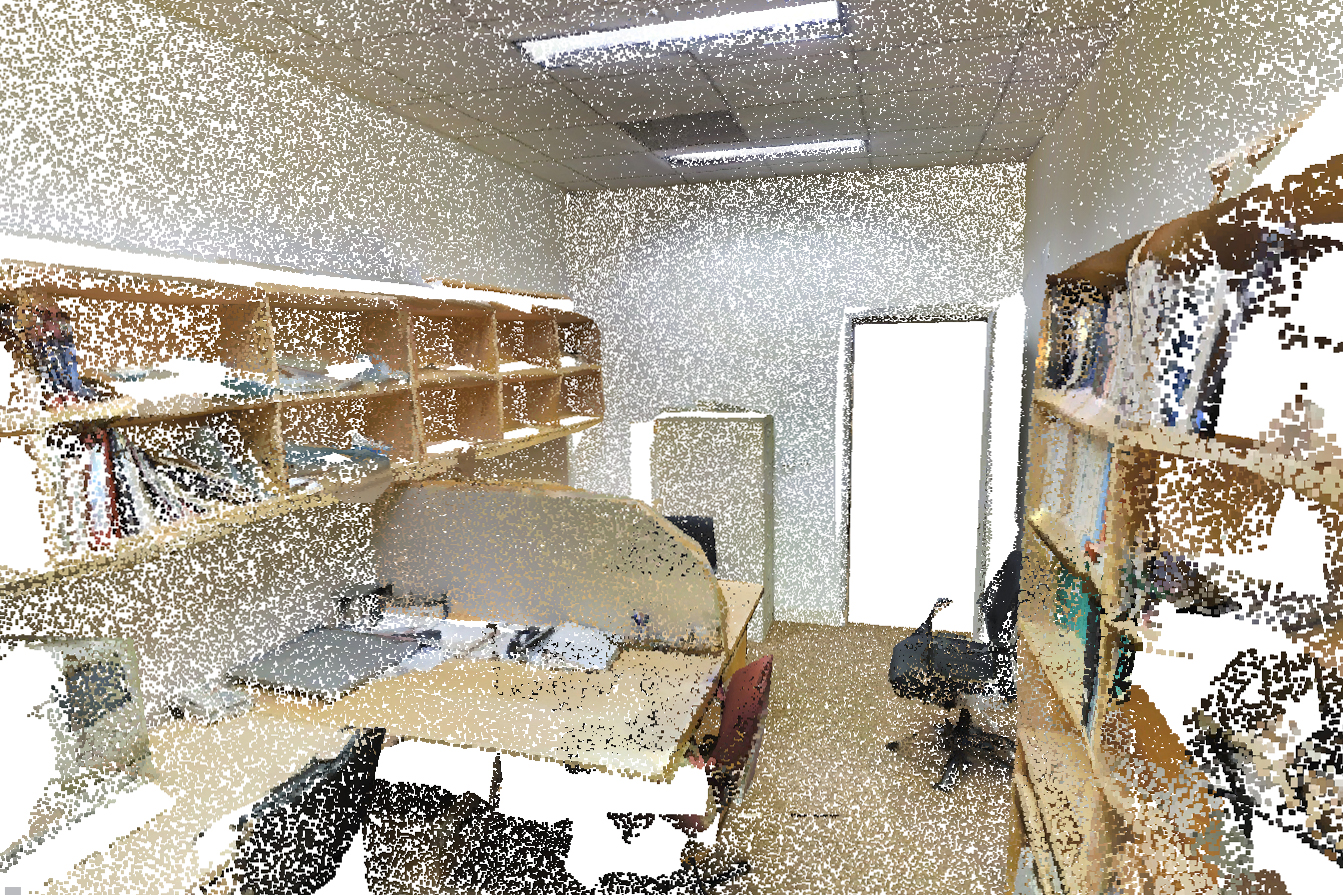}}
	\subfloat{\includegraphics[width=.3\linewidth]{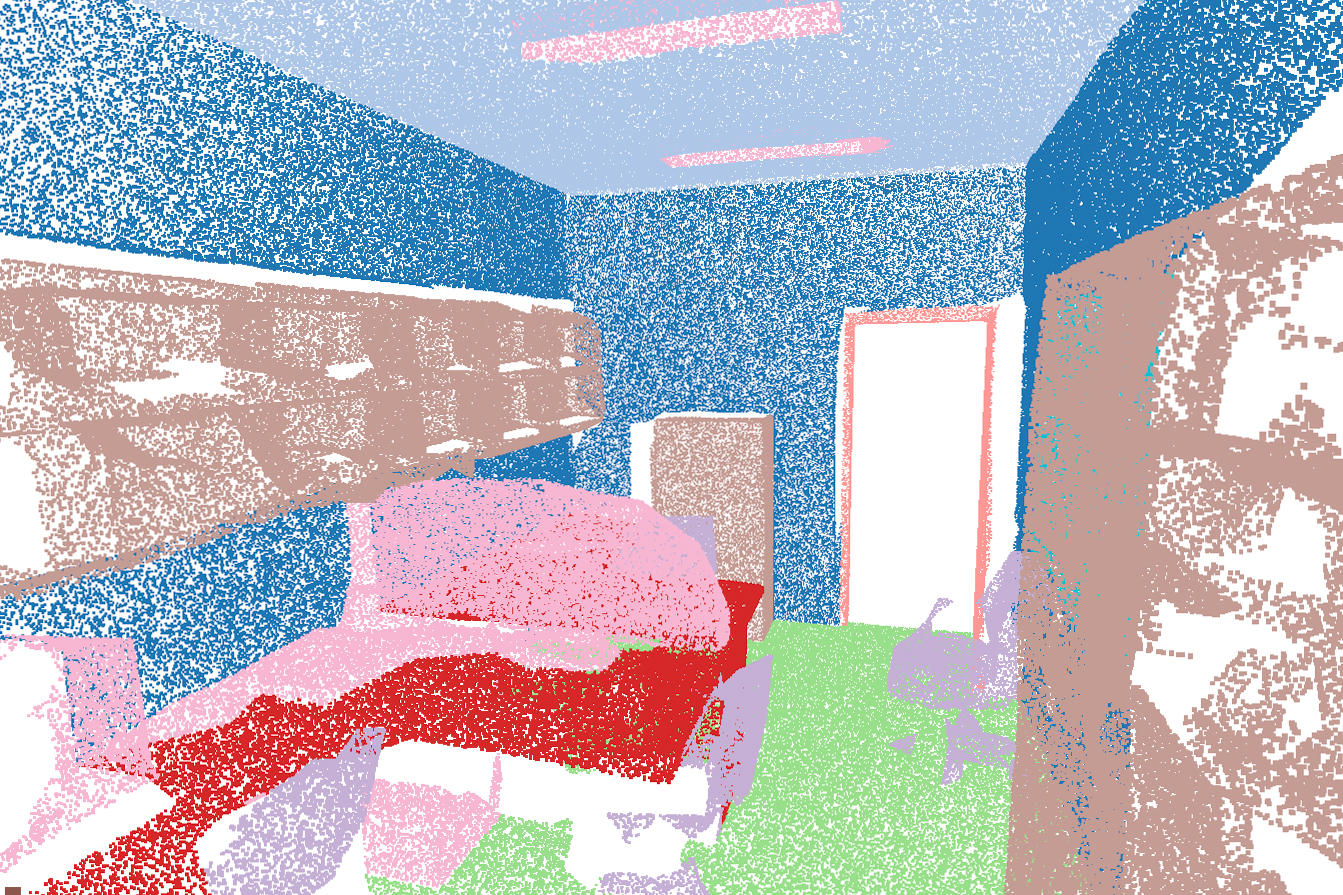}}
        \subfloat{\includegraphics[width=.3\linewidth]{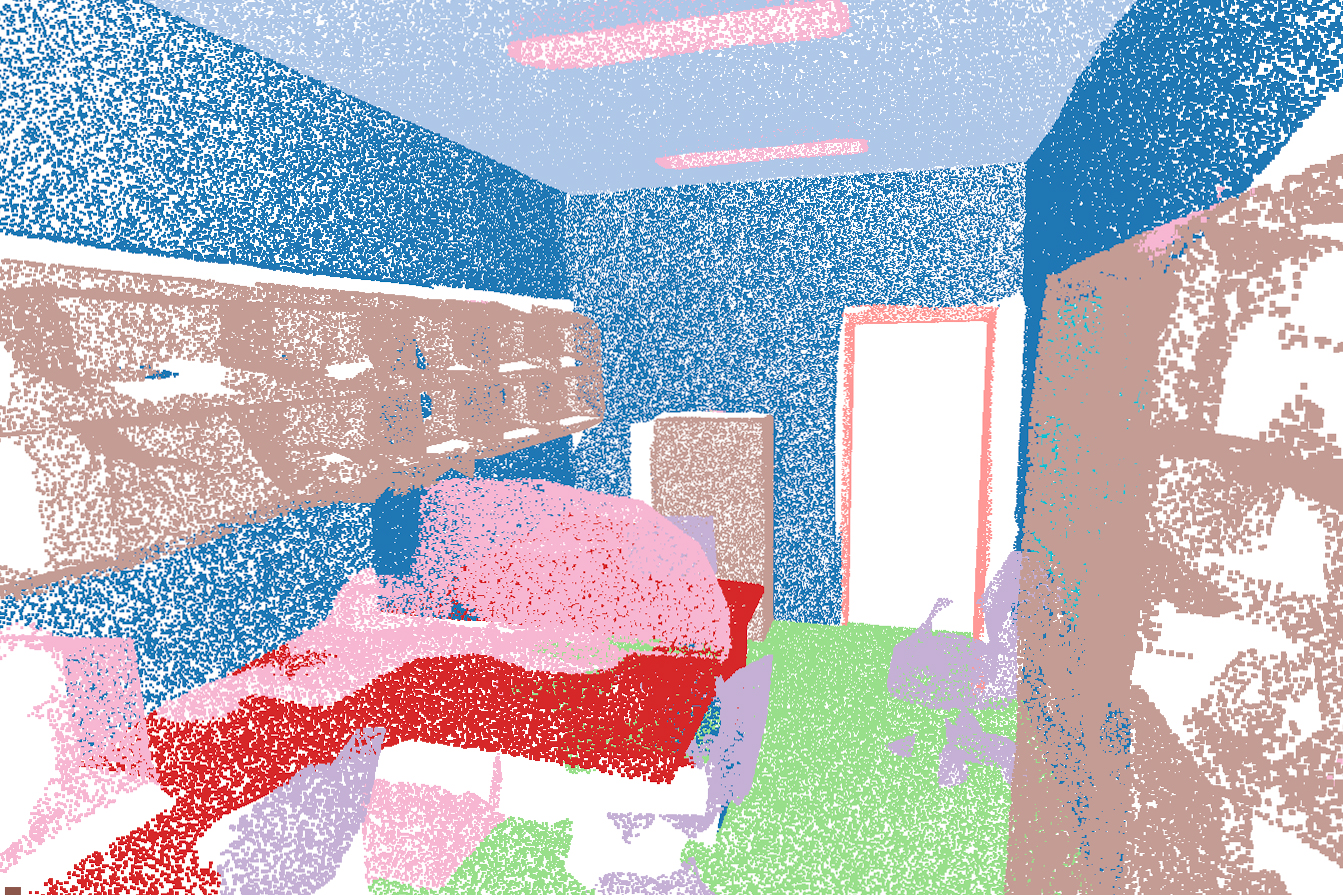}} \\
        \subfloat[Point Clouds]{\includegraphics[width=.3\linewidth]{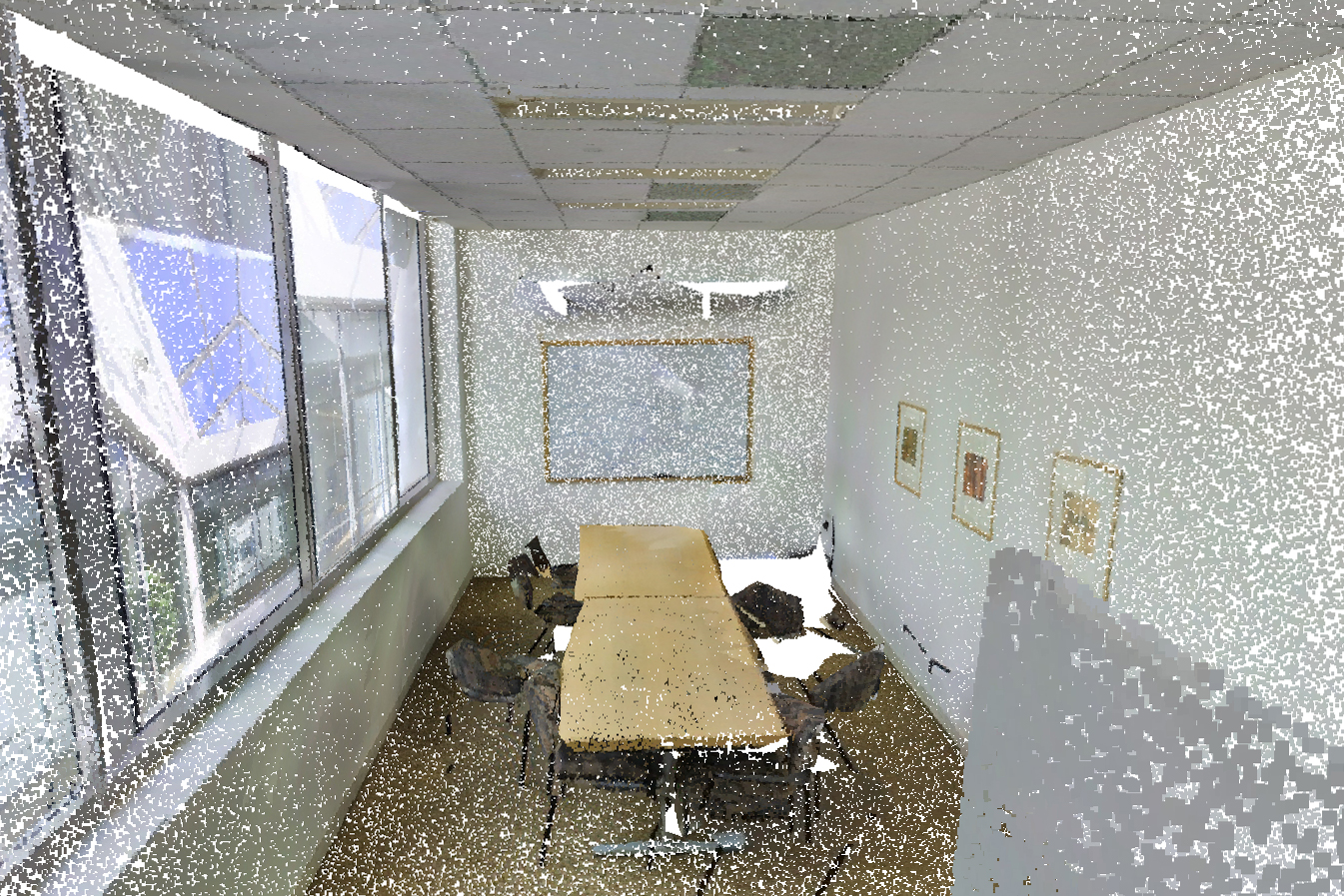}}
	\subfloat[Ground truth]{\includegraphics[width=.3\linewidth]{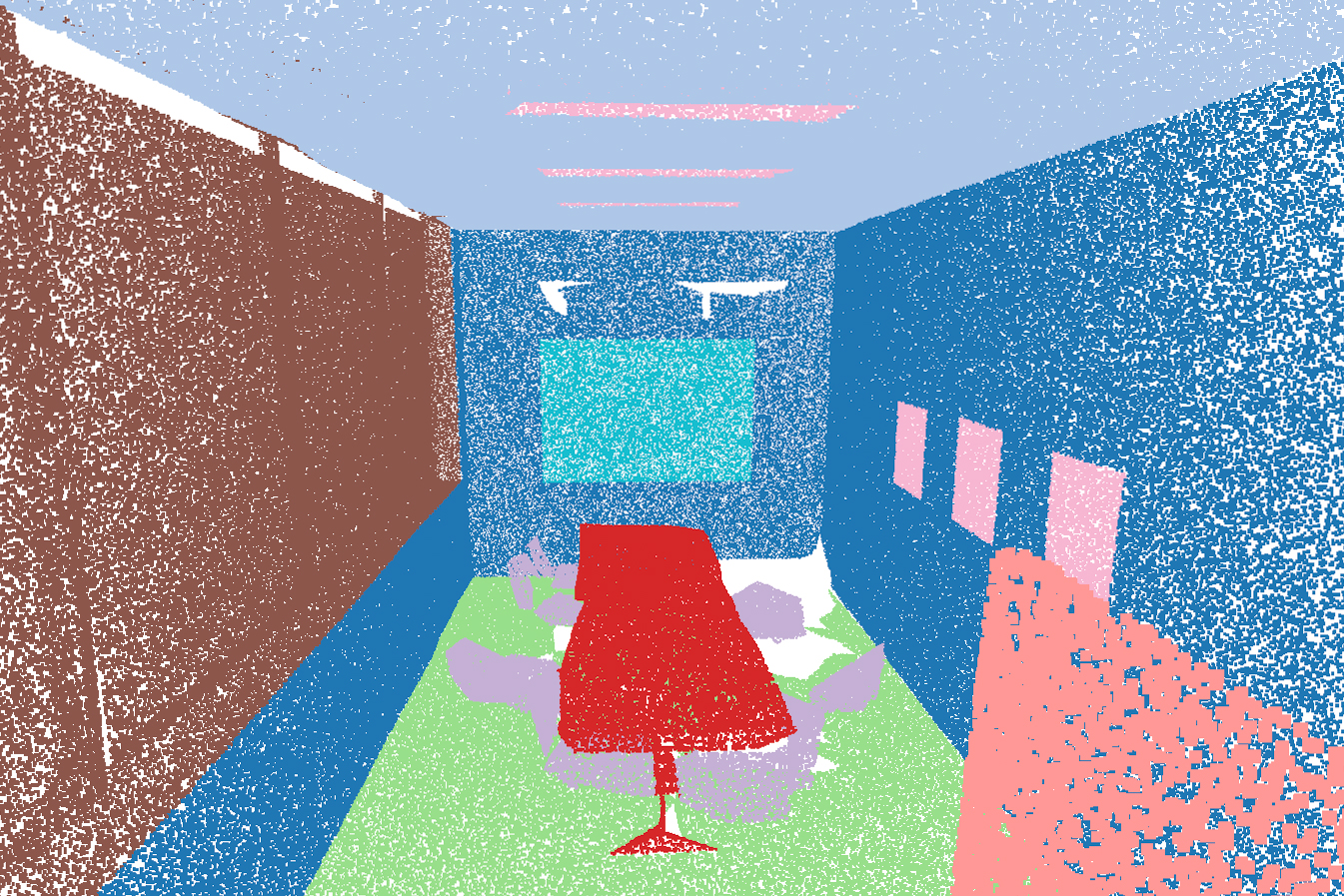}}
        \subfloat[Prediction]{\includegraphics[width=.3\linewidth]{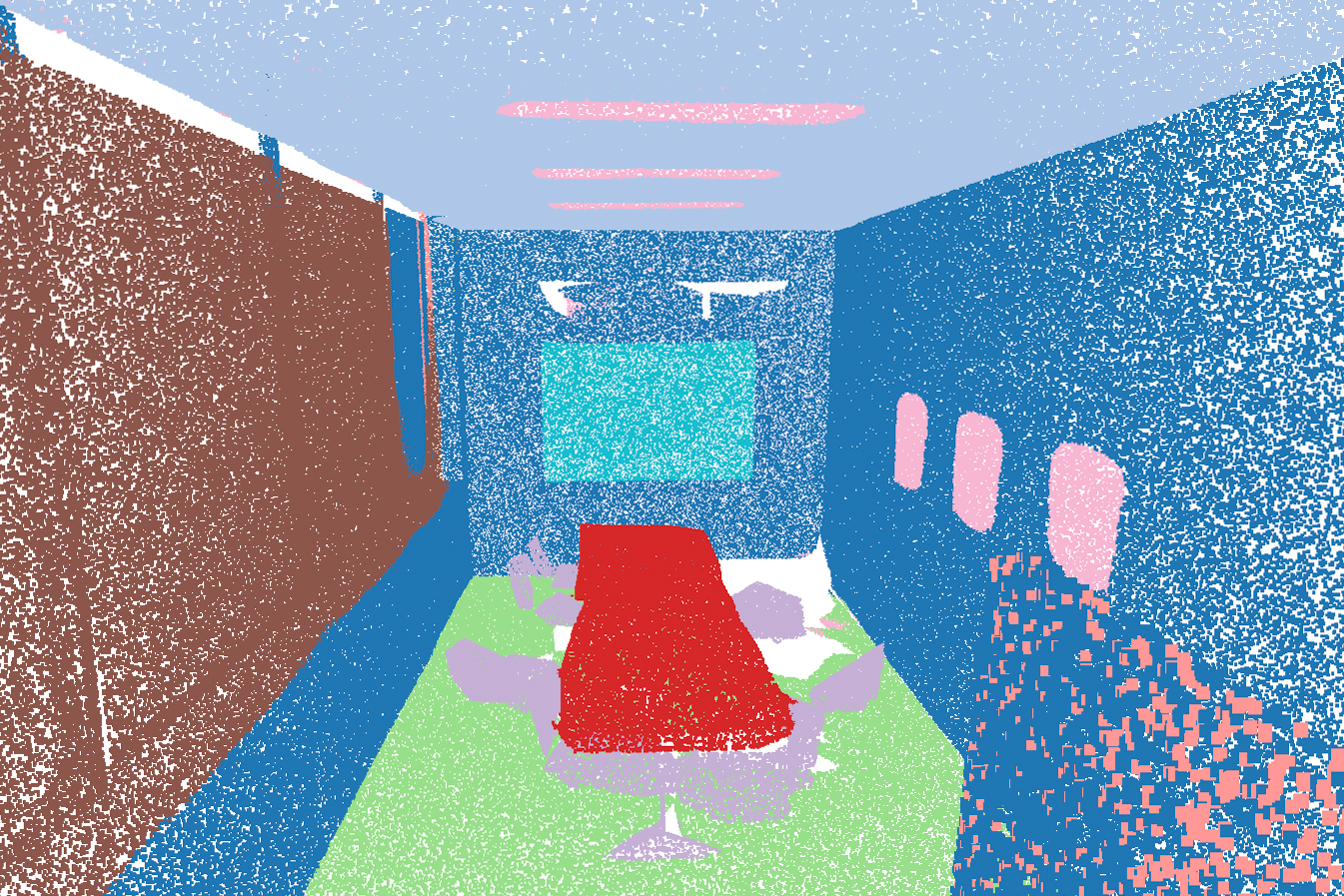}} \\
        
        \caption{Examples of semantic segmentation results on S3DIS testing set (Area 5).}
        \label{result_s3dis} 
\end{figure}

This section qualitatively presents the results of our model on ScanNetV2 and S3DIS datasets, as shown in Figures \ref{result_scannet} and \ref{result_s3dis}. Compared to ground truth, the model is capable of accurately generating semantic segmentation masks. However, there is a possibility of erroneously grouping small objects with nearby ones, or encountering some ambiguity at object boundaries.

\section{Conclusion}
In this paper, we introduce an innovative approach for 3D weakly supervised semantic segmentation by integrating 2D images. Our method aims to effectively augment the sparse labeling of 3D point clouds by leveraging the geometric correspondence between 2D views and 3D point clouds. This involves utilizing segmentation masks derived from 2D foundational models and back-projecting them to 3D space. By propagating the initial limited annotations onto the 3D masks, we substantially increase the available labels. Additionally, we incorporate consistency regularization and select reliable pseudo labels, which are then proportionally expanded into 3D masks, maximizing their utility. To address noise in the expanded labels, we employ noise-robust techniques to enhance model performance. Experiments on ScanNetV2 and S3DIS datasets demonstrate state-of-the-art performance. 

For future works, there is still room for improvement in the model's performance because issues like unclear boundaries and insensitivity to small objects persist. Besides, broader applications of 3D masks can be explored to maximize their effectiveness. Exploring alternative methods for handling label noise is also necessary. In the future, our method could be further explored for applications in various fields, including outdoor datasets, such as those collected by autonomous driving systems, where images and point clouds are acquired simultaneously.


\bibliographystyle{IEEEtran}
\bibliography{ref}

\newpage

\vfill

\end{document}